\pdfoutput=1

\documentclass[11pt]{article}

\usepackage{ACL2023}

\usepackage{times}
\usepackage{latexsym}

\usepackage[T1]{fontenc}

\usepackage[utf8]{inputenc}

\usepackage{microtype}

\usepackage{inconsolata}
\usepackage{amsmath}
\usepackage{hyperref}
\usepackage{url}
\usepackage{graphicx}
\usepackage{subfig}
\usepackage{multirow}
\usepackage{makecell}
\usepackage{booktabs}
\usepackage{amssymb}
\usepackage{color}
\usepackage{adjustbox}

\newcommand{\gmodel}{RankCSE~}{}
\newcommand{\gmodelnospace}{RankCSE}{}

\title{\gmodel: Unsupervised Sentence Representation Learning via \\ Learning to Rank}

\author{ \textbf{Jiduan Liu\textsuperscript{1,2}\thanks{\llap{}\:\:\:Work done during internship at Meituan.}\ \ , Jiahao Liu\textsuperscript{3}, Qifan Wang\textsuperscript{4}, Jingang Wang\textsuperscript{3}, Wei Wu\textsuperscript{3} }\\
\textbf{Yunsen Xian\textsuperscript{3}, Dongyan Zhao\textsuperscript{1,2,5,6}$^\dagger$, Kai Chen\textsuperscript{7}, Rui Yan\textsuperscript{8,9}}\thanks{$^\dagger$ Corresponding authors: Dongyan Zhao
(zhaody@pku.edu.cn) and Rui Yan (ruiyan@ruc.edu.cn).} \\
\textsuperscript{1}Wangxuan Institute of Computer Technology, Peking University \\
\textsuperscript{2}Center for Data Science, AAIS, Peking University; \textsuperscript{3}Meituan; \textsuperscript{4}Meta AI \\
\textsuperscript{5}National Key Laboratory of General Artificial Intelligence \\
\textsuperscript{6}BIGAI, Beijing, China; \textsuperscript{7}School of Economics, Peking University\\
\textsuperscript{8}Gaoling School of Artificial Intelligence, Renmin University of China \\
\textsuperscript{9}Engineering Research Center of \\Next-Generation Intelligent Search and Recommendation, Ministry of Education \\
\texttt{\{liujiduan,chen.kai,zhaody\}@pku.edu.cn}, \texttt{ruiyan@ruc.edu.cn}, \texttt{wqfcr@fb.com} \\
\texttt{\{liujiahao12,wangjingang02,xianyunsen\}@meituan.com}, \texttt{wuwei19850318@gmail.com}}

\begin{document}
\maketitle
\begin{abstract}
Unsupervised sentence representation learning is one of the fundamental problems in natural language processing with various downstream applications. Recently, contrastive learning has been widely adopted which derives high-quality sentence representations by pulling similar semantics closer and pushing dissimilar ones away. 
However, these methods fail to capture the fine-grained ranking information among the sentences, where each sentence is only treated as either positive or negative. In many real-world scenarios, one needs to distinguish and rank the sentences based on their similarities to a query sentence, e.g., very relevant, moderate relevant, less relevant, irrelevant, etc.  
In this paper, we propose a novel approach, \gmodelnospace, for unsupervised sentence representation learning, which incorporates ranking consistency and ranking distillation with contrastive learning into a unified framework. In particular, we learn semantically discriminative sentence representations by simultaneously ensuring ranking consistency between two representations with different dropout masks, and distilling listwise ranking knowledge from the teacher.
An extensive set of experiments are conducted on both semantic textual similarity (STS) and transfer (TR) tasks. Experimental results demonstrate the superior performance of our approach over several state-of-the-art baselines.
\end{abstract} 

\begin{table}[t]
\begin{adjustbox}{width=0.98\columnwidth,center}
\tiny
\tabcolsep=0.1cm
\begin{tabular}{p{3.2cm}ccc}
\toprule
\makecell[c]{\bf Target Sentences} & \bf Label & \bf SimCSE & \bf \gmodel \\
\midrule
$\bullet$ A woman is breaking eggs & 4.80 (1) & 0.93 (2) & 0.97 (1)\\
$\bullet$ A man is cracking eggs & 3.60 (2) & 0.94 (1) & 0.91 (2) \\
$\bullet$ A woman is talking to a man & 1.60 (3) & 0.45 (5) & 0.65 (3) \\
$\bullet$ A man and a woman are speaking & 1.40 (4) & 0.47 (3) & 0.61 (4)\\
$\bullet$ A man is talking to a boy & 1.00 (5) & 0.46 (4) & 0.56 (5) \\
\midrule
\multicolumn{4}{l}{\textbf{Query Sentence:} A woman is cracking eggs} \\
\midrule
\midrule
$\bullet$ Broccoli are being cut by a woman & 4.80 (1) & 0.82 (2) & 0.95 (1)\\
$\bullet$ A woman is slicing vegetables & 4.20 (2) & 0.83 (1) & 0.91 (2) \\
$\bullet$ A woman is cutting some plants & 3.50 (3) & 0.74 (5) & 0.87 (3) \\
$\bullet$ There is no woman cutting broccoli & 3.40 (4) & 0.76 (3) & 0.85 (4)\\
$\bullet$ A woman is cutting some flowers & 2.87 (5) & 0.71 (7) & 0.81 (5) \\
$\bullet$ A man is slicing tomatoes & 2.60 (6) & 0.75 (4) & 0.79 (6) \\
$\bullet$ A man is cutting tomatoes & 2.40 (7) & 0.73 (6) & 0.76 (7) \\
\midrule
\multicolumn{4}{l}{\textbf{Query Sentence:} A woman is cutting broccoli} \\
\bottomrule
\end{tabular}
\end{adjustbox}
\captionof{table}{Two examples of a query sentence and several target sentences from the STS datasets, with their similarity scores and rankings. The label scores are from human annotations. The SimCSE \cite{DBLP:conf/emnlp/GaoYC21} and \gmodel similarity scores are from the model predictions respectively, with the corresponding ranking positions. It can be seen that sentence rankings based on SimCSE are incorrect, while \gmodel generates more effective scores with accurate rankings.}
\vspace{-3mm}
\label{tab:case}
\end{table}
\begin{figure}[t]
\centering
\subfloat[KCC]{\includegraphics[width=0.49\columnwidth]{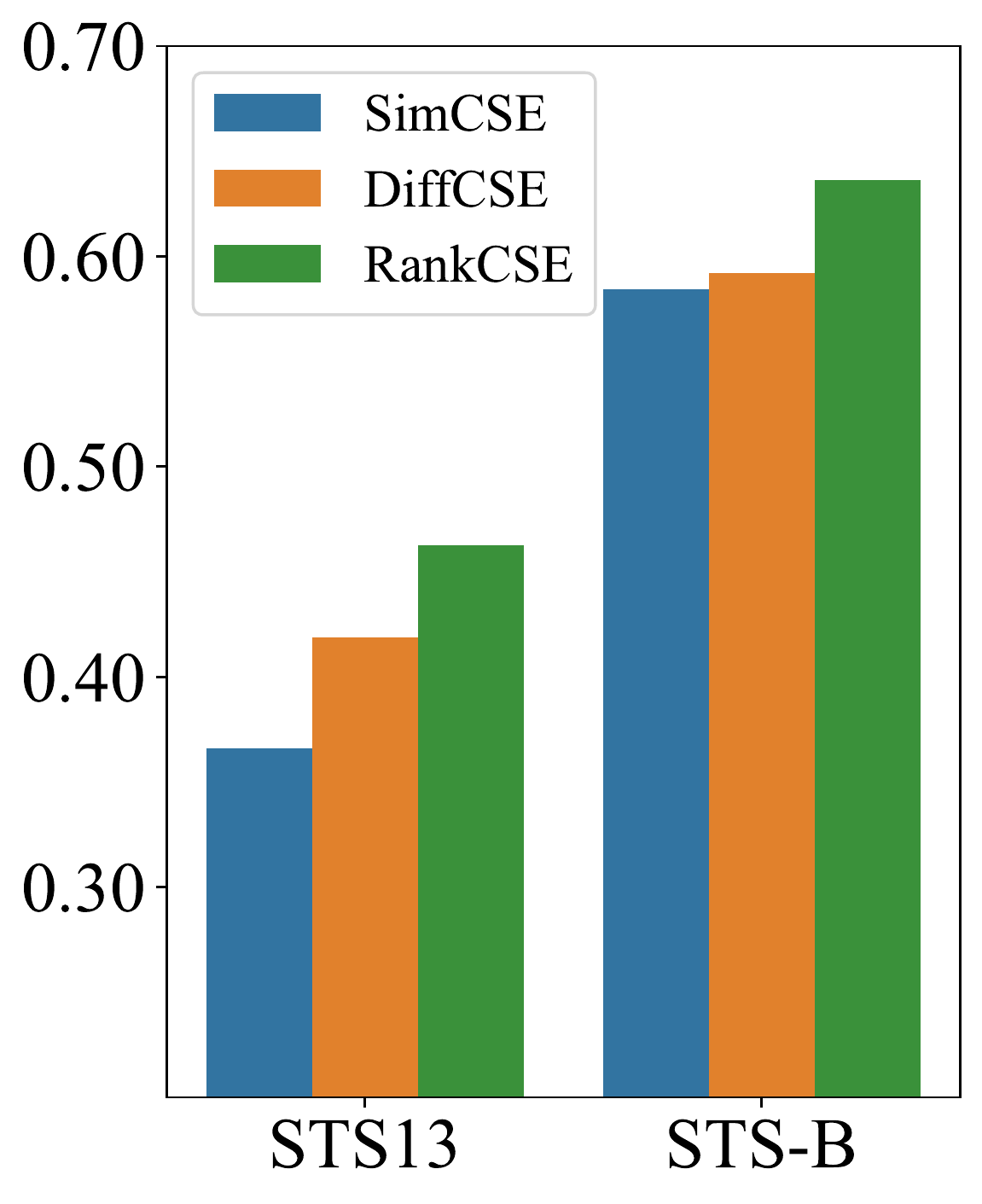}\label{fig:kcc}}
\subfloat[NDCG]{\includegraphics[width=0.49\columnwidth]{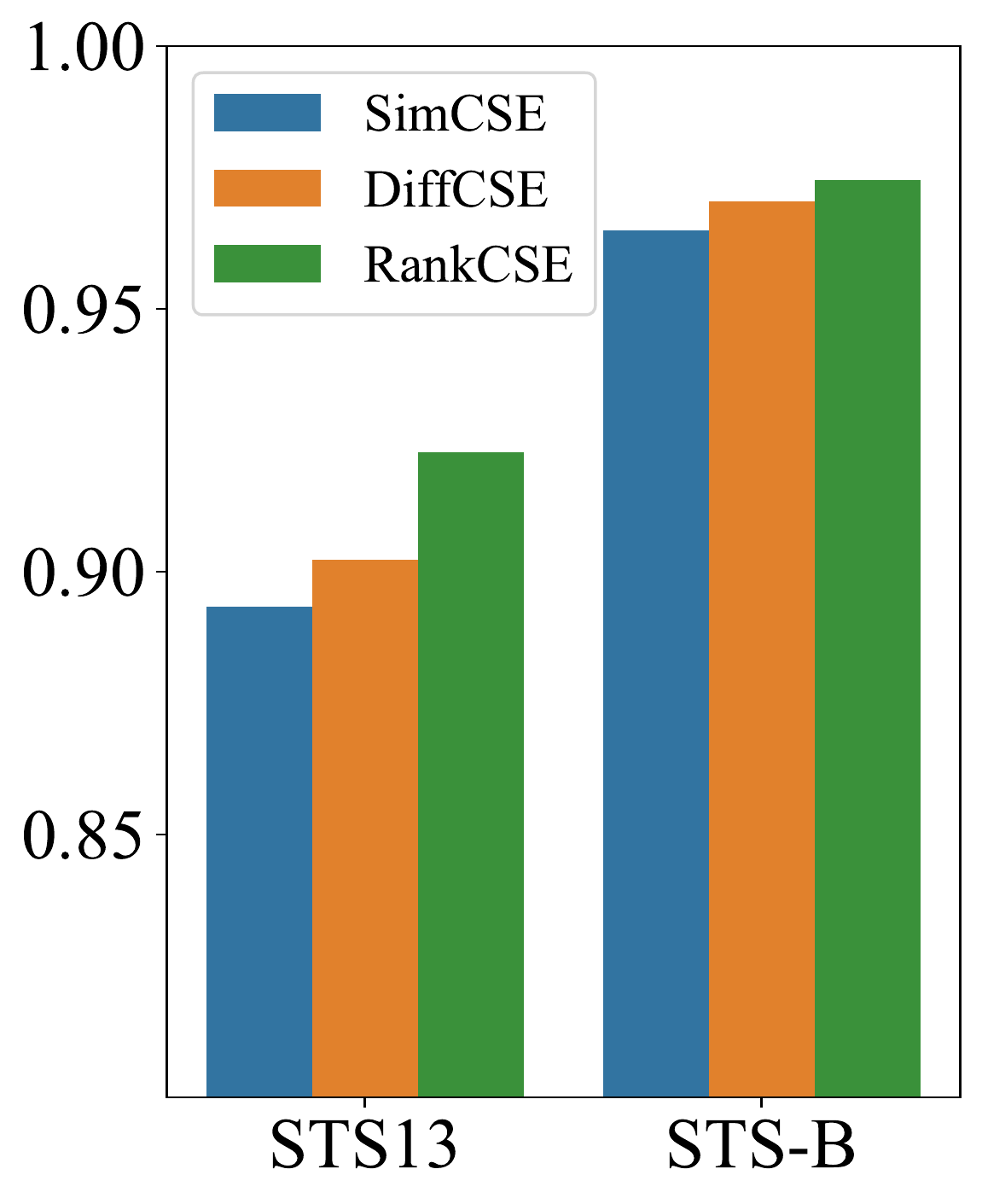}\label{fig:ndcg}}
\captionof{figure}{Sentence representation performances on ranking metrics KCC and NDCG (detailed in Appendix \ref{ap:rank}). It can be seen that \gmodel captures more fine-grained ranking information than SimCSE \cite{DBLP:conf/emnlp/GaoYC21} and DiffCSE \cite{DBLP:conf/naacl/ChuangDLZCS0YKG22}.}
\label{fig:ranking_metrics}
\vspace{-3mm}
\end{figure}
\section{Introduction}
Sentence representation learning refers to the task of encoding sentences into fixed-dimensional embeddings. The sentence embeddings can be leveraged in various applications, including information retrieval \cite{DBLP:conf/icml/LeM14}, text clustering \cite{DBLP:conf/asist/MaZH16} and semantic textual similarity comparison \cite{DBLP:conf/semeval/AgirreCDG12}. With the recent success of pre-trained language models (PLMs), such as BERT/RoBERTa \cite{DBLP:conf/naacl/DevlinCLT19,DBLP:journals/corr/abs-1907-11692}, a straightforward way to generate sentence representations is to directly use the [CLS] token embedding or the average token embeddings from the last layer of PLMs \cite{DBLP:conf/emnlp/ReimersG19}. However, several studies \cite{DBLP:conf/emnlp/Ethayarajh19,DBLP:conf/emnlp/LiZHWYL20} have found that the native sentence representations derived by PLMs occupy a narrow cone in the vector space, and thus severely limits their representation capabilities, which is known as the \emph{anisotropy} problem.

Supervised methods like SBERT \cite{DBLP:conf/emnlp/ReimersG19} usually generate better sentence representations, but require fine-tuning on a large amount of labeled data. Recent unsupervised models \cite{DBLP:conf/iclr/CarlssonGGHS21,DBLP:conf/acl/0004HLB020,DBLP:conf/acl/GiorgiNWB20} adopt contrastive learning framework without any labels, which pulls similar semantics closer and pushes dissimilar ones away. These methods usually design different augmentation algorithms for generating positive examples, such as back-translation \cite{DBLP:conf/acl/0004HLB020}, dropout \cite{DBLP:conf/emnlp/GaoYC21} and token shuffling or cutoff \cite{DBLP:conf/acl/YanLWZWX20}. In-batch negatives are further combined with the positives. 
Despite achieving promising results, they treat positives/negatives equally without capturing the fine-grained semantic ranking information, resulting in less effective sentence representations which fail to distinguish between very similar and less similar sentences. For example, Table \ref{tab:case} shows two examples of a query sentence and several target sentences from semantic textual similarity datasets. It is clear that the similarity scores produced by the contrastive learning method SimCSE are not optimized, where the sentence rankings are not preserved in the learned representations. On the other hand, our RankCSE generates effective sentence representations with consistent rankings to the ground-truth labels. Figure \ref{fig:ranking_metrics} further shows the advantage of RankCSE in terms of two ranking metrics. The fine-grained ranking information is crucial in various real-world applications including search and recommendation. The ability to differentiate between subtle distinctions in sentence meaning can help these systems provide more relevant and accurate results, leading to a better user experience.
Therefore, it is an important problem to learn ranking preserving sentence representations from unsupervised data.



To obtain semantically discriminative sentence representations, we propose a novel approach, \gmodelnospace, which incorporates ranking consistency and ranking distillation with contrastive learning into a unified framework. Specifically, our model ensures ranking consistency between two representations with different dropout masks and minimizes the Jensen-Shannon (JS) divergence as the learning objective. In the meanwhile, our model also distills listwise ranking knowledge from the teacher model to the learned sentence representations. In our work, we explore two listwise ranking methods, ListNet \cite{DBLP:conf/icml/CaoQLTL07} and ListMLE \cite{DBLP:conf/icml/XiaLWZL08}, and utilize the pre-trained SimCSE \cite{DBLP:conf/emnlp/GaoYC21} models with coarse-grained semantic ranking information as the teachers to provide pseudo ranking labels. Our \gmodel is able to generalize fine-grained ranking information from the weak ranking knowledge learned by SimCSE.
We conduct an extensive set of experiments on semantic textual similarity (STS) and transfer (TR) tasks. Experimental results show that \gmodel outperforms the existing state-of-the-art baselines.

\begin{figure*}
    \centering \includegraphics[width=1.95\columnwidth]{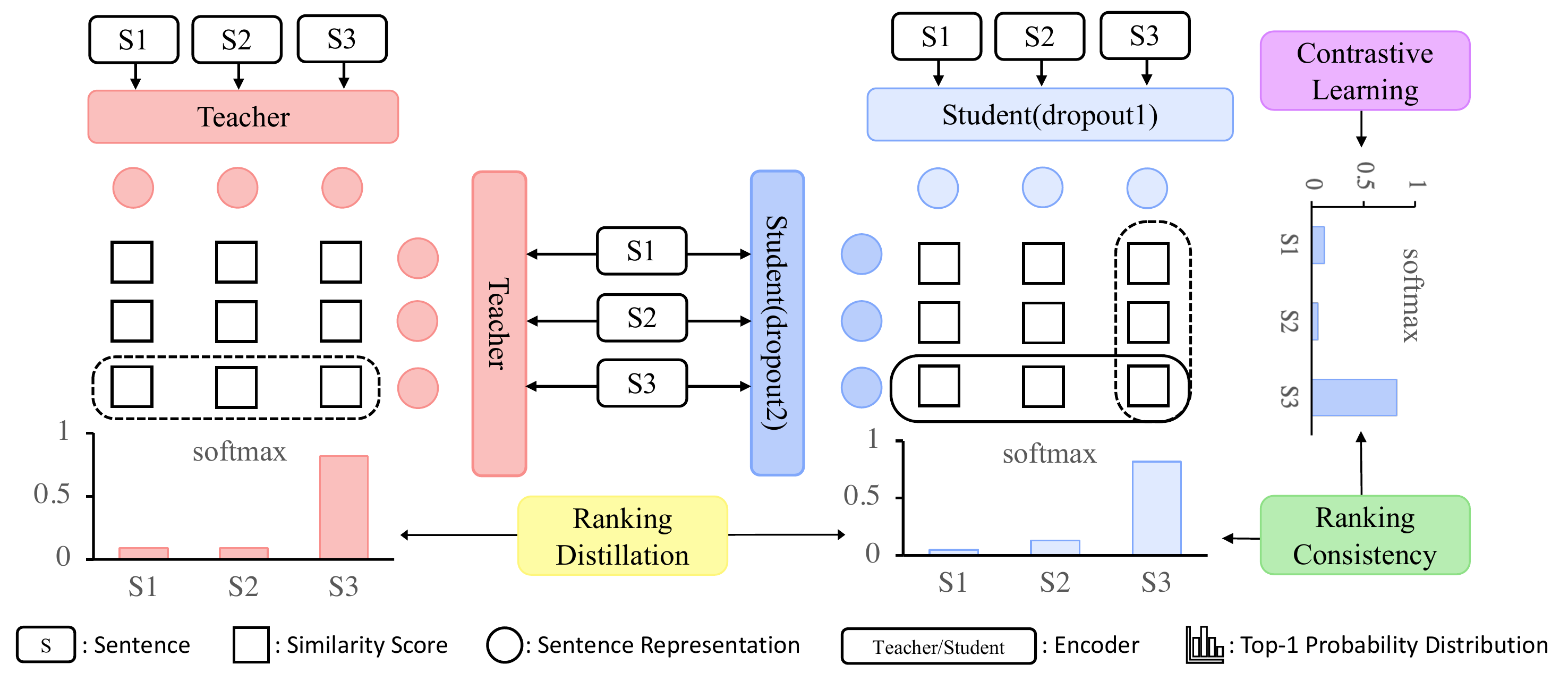}
    \caption{The framework of \gmodel which consists of three components: (1) contrastive learning object; (2) ranking consistency loss which ensures ranking consistency between two representations with different dropout masks; (3) ranking distillation loss which distills listwise ranking knowledge from the teacher.}
    \label{fig:model}
    \vspace{-3mm}
\end{figure*}
\section{Related Work}

\paragraph{Unsupervised Sentence Representation Learning} Early works typically augment the idea of word2vec \cite{DBLP:conf/nips/MikolovSCCD13} to learn sentence representations, including Skip-Thought \cite{DBLP:conf/nips/KirosZSZUTF15}, FastSent \cite{DBLP:conf/naacl/HillCK16} and Quick-Thought \cite{DBLP:conf/iclr/LogeswaranL18}. With the great success of PLMs, various attempts focus on generating sentence representations by leveraging the embedding of [CLS] token or applying mean pooling on the last layer of BERT \cite{DBLP:conf/emnlp/ReimersG19}. However, \citet{DBLP:conf/emnlp/Ethayarajh19} identifies the \emph{anisotropy} problem in language representations, which means the native learned embeddings from PLMs occupy a narrow cone in the vector space. BERT-flow \cite{DBLP:conf/emnlp/LiZHWYL20} and BERT-whitening \cite{DBLP:journals/corr/abs-2103-15316} propose to resolve the \emph{anisotropy} problem through post-processing.

Recently, contrastive learning has been adopted to learn sentence representations by designing different augmentation methods \cite{DBLP:conf/emnlp/ZhangHLLB20,DBLP:conf/iclr/CarlssonGGHS21,DBLP:conf/acl/GiorgiNWB20,DBLP:conf/acl/YanLWZWX20,DBLP:conf/acl/KimYL20,DBLP:conf/emnlp/GaoYC21}. A typical example SimCSE uses dropout as a data augmentation strategy and is also the foundation of many following works. ArcCSE \cite{DBLP:conf/acl/ZhangZWXLZ22} enhances the pairwise discriminative power and models the entailment relation among triplet sentences. 
DCLR \cite{DBLP:conf/acl/ZhouZZW22} alleviates the influence of improper negatives. DiffCSE \cite{DBLP:conf/naacl/ChuangDLZCS0YKG22} introduces equivariant contrastive learning to SimCSE. PCL \cite{DBLP:journals/corr/abs-2201-12093} proposes contrastive representation learning with diverse augmentation strategies for an inherent anti-bias ability. InfoCSE \cite{DBLP:journals/corr/abs-2210-06432} learns sentence representations with the ability to reconstruct the original sentence fragments. Generative learning techniques \cite{DBLP:conf/emnlp/Wang0G21,DBLP:conf/emnlp/WuZ22} have also been proposed to enhance the linguistic interpretability of sentence representations.
Although achieving promising results, these methods fail to capture the fine-grained ranking knowledge among the sentences.

\paragraph{Learning to Rank} 
Given a query example, learning to rank aims to rank a list of examples according to their similarities with the query. Learning to rank methods can be divided into three categories: pointwise \cite{DBLP:conf/nips/LiBW07}, pairwise \cite{DBLP:conf/icml/BurgesSRLDHH05,DBLP:conf/nips/BurgesRL06} and listwise \cite{DBLP:conf/icml/CaoQLTL07,DBLP:conf/icml/XiaLWZL08,DBLP:conf/icml/VolkovsZ09,DBLP:journals/corr/abs-2102-07831}. Pointwise methods optimize the similarity between the query and each example, while pairwise approaches learn to correctly model the preference between two examples. Listwise methods directly evaluate the ranking of a list of examples based on the ground truth. In our framework, we leverage listwise ranking objectives for learning effective sentence representations, which have shown better performance compared to pointwise and pairwise methods.

\section{Preliminary}
We provide some conceptual explanations and definitions in learning to rank.

\paragraph{Top One Probability} Given the scores of all objects $S= \{s_i\}_{i=1}^n$, the top one probability of an object is the probability of its being ranked at top-1: $\widetilde{s_i} = \frac{e^{s_i/\tau}}{\sum_{j=1}^ne^{s_j/\tau}}$ where $\tau$ is a temperature hyperparameter, usually utilized to smooth the distribution. We simply denote the formulation for calculating the top one distribution based on the scores $S$ as: $\widetilde{S_{\tau}} = \sigma(S/\tau)$.

\paragraph{Permutation Probability} Let $\pi=\{\pi(i)\}_{i=1}^n$ denote a permutation of the object indexes, which represents that the $\pi(i)$-th sample is ranked i-th. The probability of a specific permutation $\pi$ is given as: $P(\pi|S,\tau) = \prod_{i=1}^n \frac{e^{s_{\pi(i)}/\tau}}{\sum_{j=i}^n e^{s_{\pi(j)}/\tau}}$.
\section{Methodology}

\subsection{Problem Formulation}
Our goal is to learn sentence representations such that semantic similar sentences stay close while dissimilar ones should be far away in an unsupervised manner. Specifically, We aim to find an optimal function $f$ that maps a sentence $s \in p_{\rm s}$ to a $d$-dimensional vector $f(s) \in p_{\rm e} \subseteq \mathcal{R}^d$, where $p_{\rm s}$ and $p_{\rm e}$ denote the distributions of sentences and sentence representations, respectively. Supposing $s_1$ and $s_2$ are more semantic similar than $s_1$ and $s_3$ ($s_1,s_2,s_3 \in p_{\rm s}$), a good mapping function $f$ should satisfy that the distance between $f(s_1)$ and $f(s_2)$ is smaller than that between $f(s_1)$ and $f(s_3)$, i.e., $d(f(s_1),f(s_2)) < d(f(s_1),f(s_3))$, where $d$ is the distance metric such as Euclidean distance and cosine distance. In this way, the similarities among the sentences are preserved in the learned sentence representations.

The general idea of \gmodel is to learn semantically discriminative sentence representations by capturing the ranking information among the sentences. As shown in Figure \ref{fig:model}, our model consists of three components: (1) contrastive learning objective (Section \ref{sec3.1}); (2) ranking consistency loss which ensures ranking consistency between two representations with different dropout masks (Section \ref{sec3.2}); (3) ranking distillation loss which distills listwise ranking knowledge from the teacher (Section \ref{sec3.3}).

\subsection{Contrastive Learning} \label{sec3.1}
Contrastive learning aims to learn effective representations by pulling similar semantics closer and pushing away dissimilar ones. SimCSE \cite{DBLP:conf/emnlp/GaoYC21} creates positive examples by applying different dropout masks and takes a cross-entropy object with in-batch negatives \cite{DBLP:conf/kdd/ChenSSH17}. More specifically, for any sentence $x_i$ in a min-batch, we send it to the encoder $f(\cdot)$ twice and obtain two representations with different dropout masks $f(x_i)$, $f(x_i)'$. SimCSE use the InfoNCE loss \cite{DBLP:journals/corr/abs-1807-03748} as the training objective:
\begin{equation} \small \label{eq:contrastive}
    \mathcal{L}_{\rm{infoNCE}} = - \sum_{i=1}^N\log \frac{e^{\phi(f(x_i),f(x_i)')/\tau_1}}{\sum_{j=1}^Ne^{\phi(f(x_i),f(x_j)')/\tau_1}},
\end{equation}
where $N$ is the batch size, $\tau_1$ is a temperature hyperparameter and $\phi(f(x_i),f(x_j)')=\frac{f(x_i)\top f(x_j)'}{\Vert f(x_i) \Vert \cdot \Vert f(x_j)' \Vert}$ is the cosine similarity used in this work. Essentially, the contrastive learning objective is equivalent to maximizing the top one probability of the positive sample.

Although contrastive learning is effective in separating positive sentences with negative ones, it ignores the continuity modeling of the similarity. In other words, it is not effective in distinguishing highly similar sentences with moderate similar ones.
To address this issue, we propose to directly model the ranking information among the sentences, which could enhance the discrimination of semantic similarity in the learned sentence representations.

\subsection{Ranking Consistency} \label{sec3.2}
The main drawback of contrastive learning is that the distinction between the in-batch negatives is not modeled, resulting in less effective sentence representations in capturing the fine-grained sentence similarity. Therefore, instead of treating the negatives equivalently, we propose to explicitly model the ranking information within the sentences by ensuring the ranking consistency between the two similarity sets (circled by the solid and dashed curves respectively in the right part of Figure \ref{fig:model}). 

Concretely, by taking a close look at the contrastive modeling in Section \ref{sec3.1}, there are two sets of sentence representations, $f(x_i)$ and $f(x_i)'$, derived from different dropout masks. For each sentence $x_i$, two lists of similarities with other sentences can be naturally obtained from the two representations, i.e., $S(x_i)=\{\phi(f(x_i),f(x_j)')\}_{j=1}^N$ and $S(x_i)'=\{\phi(f(x_i)',f(x_j))\}_{j=1}^N$. We then enforce the ranking consistency between these two similarity lists in our modeling. Intuitively, all corresponding elements in $S(x_i)$ and $S(x_i)'$ should have the same ranking positions.


Given two similarity lists $S(x_i)$ and $S(x_i)'$, we can obtain their top one probability distributions $\widetilde{S}_{\tau_1}(x_i)=\sigma(S(x_i)/\tau_1)$, $\widetilde{S}_{\tau_1}(x_i)'=\sigma(S(x_i)'/\tau_1)$. The ranking consistency can be ensured by minimizing the Jensen-Shannon (JS) divergence between the two top one probability distributions:
\begin{equation} \label{eq:consistency}
    \small
    \begin{aligned}
    &\mathcal{L}_{\rm{consistency}} = \sum_{i=1}^N \rm{JS}(P_i||Q_i) \\
    &= \frac{1}{2}\sum_{i=1}^N(\rm{KL}(P_i||\frac{P_i+Q_i}{2})+\rm{KL}(Q_i||\frac{P_i+Q_i}{2})) \\
    &= \frac{1}{2}\sum_{i=1}^N (P_i\log (\frac{2P_i}{P_i+Q_i})+Q_i\log (\frac{2Q_i}{P_i+Q_i})),
    \end{aligned}
\end{equation} where $P_i$ and $Q_i$ represents $\widetilde{S}_{\tau_1}(x_i)$ and $\widetilde{S}_{\tau_1}(x_i)'$ respectively. The reason we choose JS divergence instead of Kullback-Leibler (KL) divergence is that the two distributions are symmetric rather than one side being the ground truth.

\subsection{Ranking Distillation} \label{sec3.3}
Contrastive learning based methods like SimCSE learn effective sentence representations with coarse-grained semantic ranking information (shown in Appendix \ref{ap:case} and \ref{ap:rank}), which have demonstrated their effectiveness in various downstream tasks. Orthogonal to ranking consistency, we further introduce ranking distillation by distilling the ranking knowledge from pre-trained teacher models into our learned sentence representations, to generalize effective ranking information from the weak ranking knowledge learned in the teachers. More specifically, for each sentence in a min-batch, we obtain the similarity score list from the teacher model, which is then served as pseudo ranking labels in the ranking distillation. The intuitive idea is to transfer the ranking knowledge from the teacher to the student as guidance for learning ranking preserved sentence representations.
In the ranking distillation, ListNet \cite{DBLP:conf/icml/CaoQLTL07} and ListMLE \cite{DBLP:conf/icml/XiaLWZL08} methods are utilized. Formally they are defined as:
\begin{equation} \label{eq:rank}
    \small
    \mathcal{L}_{\rm{rank}} = \sum_{i=1}^N \rm{rank}(S(x_i), S^T(x_i)),
\end{equation}
where $S(x_i)$ and $S^T(x_i)$ are the similarity score lists obtained from the student and the teacher, respectively, $\rm{rank}(\cdot,\cdot)$ is the listwise method.

\paragraph{ListNet} The original ListNet minimizes the  cross entropy between the permutation probability distribution and the ground truth as the training objective. However, the computations will be intractable when the number of examples $n$ is large, since the number of permutations is $n!$. To reduce the computation complexity, the top one probability distribution is usually adopted as a substitute:
\begin{equation} \small \label{eq:listnet}
    \mathcal{L}_{\rm{ListNet}} = - \sum_{i=1}^N \sigma(S^T(x_i)/\tau_3)\cdot \log ( \sigma(S(x_i)/\tau_2)),
\end{equation}
where $\tau_2$ and $\tau_3$ are temperature hyperparameters.\footnote{In practice, we exclude the score of the positive pair from the list to calculate the top one distribution used in Eq.(\ref{eq:listnet}), to enhance the ranking information of negatives, because the score of the positive pair occupies most in the full top one distribution calculated by the teacher SimCSE.} 

\paragraph{ListMLE} Different from ListNet, ListMLE aims to maximize the likelihood of the ground truth permutation $\pi_i^T$ which represents the sorted indexes of the similarity scores calculated by the teacher model. The objective of ListMLE is defined as:
\begin{equation}
    \small
    \mathcal{L}_{\rm{ListMLE}} = -\sum_{i=1}^N\log P(\pi_i^T|S(x_i),\tau_2).
\end{equation}

In this work, we propose to use a multi-teacher from which more listwise ranking knowledge can be transferred and preserved. In our experiments, we utilize the weighted average similarity scores of two teachers as pseudo ranking labels: $S^T(x_i) = \alpha S^T_1(x_i) + (1 - \alpha) S_2^T(x_i)$ where $\alpha$ is a hyperparameter to balance the weight of the teachers. 

The contrastive learning loss $\mathcal{L}_{\rm{infoNCE}}$ pushes apart the representations of different sentences to maximize the representation space, while the ranking consistency loss $\mathcal{L}_{\rm{consistency}}$ and the ranking distillation loss $\mathcal{L}_{\rm{rank}}$ pull similar negatives closer, thus capturing fine-grained semantic ranking information. Combining the above three loss functions, we can obtain the overall objective:
\begin{equation}\label{eq:totloss}
    \small
    \mathcal{L}_{\rm{final}} = \mathcal{L}_{\rm{infoNCE}} + \beta  \mathcal{L}_{\rm{consistency}} + \gamma  \mathcal{L}_{\rm{rank}},
\end{equation}
where $\beta$ and $\gamma$ are hyperparameters to balance different losses.

\section{Experiment}
\begin{table*}[t]
\small
\centering
\tabcolsep=0.2cm
\begin{adjustbox}{width=1.95\columnwidth,center}
  \begin{tabular}{p{2cm}|l|cccccccc}
    \toprule
     \bf PLMs & \bf Methods  & \bf STS12 & \bf STS13 & \bf STS14 & \bf STS15 & \bf STS16 & \bf STS-B & \bf SICK-R & \bf avg. \\
    \midrule
    \multirow{2}{*}{ Non-BERT} & GloVe(avg.) & 55.14 & 70.66 & 59.73 & 68.25 & 63.66 & 58.02 & 53.76 & 61.32 \\ 
    & USE & 64.49 & 67.80 & 64.61 & 76.83 & 73.18 & 74.92 & 76.69 & 71.22 \\
    \midrule
    \multirow{13}{*}{BERT$_{\rm base}$} & first-last avg. & 39.70 & 59.38 & 49.67 & 66.03 & 66.19 & 53.87 & 62.06 & 56.70 \\
    & +flow & 58.40 & 67.10 & 60.85 & 75.16 & 71.22 & 68.66 & 64.47 & 66.55 \\
    & +whitening & 57.83 & 66.90 & 60.90 & 75.08 & 71.31 & 68.24 & 63.73 & 66.28 \\
    & +IS & 56.77 & 69.24 & 61.21 & 75.23 & 70.16 & 69.21 & 64.25 & 66.58 \\
    & +ConSERT & 64.64 & 78.49 & 69.07 & 79.72 & 75.95 & 73.97 & 67.31 & 72.74 \\
    & +SimCSE & 68.40 & 82.41 & 74.38 & 80.91 & 78.56 & 76.85 & 72.23 & 76.25 \\
    & +DCLR & 70.81 & 83.73 & 75.11 & 82.56 & 78.44 & 78.31 & 71.59 & 77.22 \\
    & +ArcCSE & 72.08 & 84.27 & 76.25 & 82.32 & 79.54 & 79.92 & 72.39 & 78.11 \\
    & +DiffCSE & 72.28 & 84.43 & 76.47 & 83.90 & 80.54 & 80.59 & 71.23 & 78.49 \\
    & +PaSeR & 70.21 & 83.88 & 73.06 & 83.87 & 77.60 & 79.19 & 65.31 & 76.16 \\
    & +PCL & 72.84 & 83.81 & 76.52 & 83.06 & 79.32 & 80.01 & 73.38 & 78.42 \\
    & \bf +\gmodelnospace$_{\rm listNet}$ & \underline{74.38} & \underline{85.97} & \underline{77.51} & \underline{84.46} & \bf 81.31 & \underline{81.46} & \bf 75.26 & \underline{80.05} \\
    & \bf +\gmodelnospace$_{\rm listMLE}$ & \bf 75.66 & \bf 86.27 & \bf 77.81 & \bf 84.74 & \underline{81.10} & \bf 81.80 & \underline{75.13} & \bf 80.36 \\
    \midrule
    \multirow{6}{*}{BERT$_{\rm large}$} 
    & +SimCSE & 70.88 & 84.16 & 76.43 & 84.50 & 79.76 & 79.26 & 73.88 & 78.41 \\ 
    & +DCLR &71.87 & 84.83 & 77.37 & 84.70 & 79.81 & 79.55 & 74.19 & 78.90 \\
    & +ArcCSE & 73.17 & 86.19 & 77.90 & 84.97 & 79.43 & 80.45 & 73.50 & 79.37 \\
    & +PCL & \underline{74.87} & 86.11 & 78.29 & \bf 85.65 & 80.52 & \bf 81.62 & 73.94 & 80.14 \\
    & \bf +\gmodelnospace$_{\rm listNet}$& 74.75 & \underline{86.46} & \underline{78.52} & 85.41 & \underline{80.62} & 81.40 & \bf 76.12 & \underline{80.47} \\
    & \bf +\gmodelnospace$_{\rm listMLE}$& \bf 75.48 & \bf 86.50 & \bf 78.60 & \underline{85.45} & \bf 81.09 & \underline{81.58} &  \underline{75.53} & \bf 80.60 \\
    \midrule
    \multirow{6}{*}{RoBERTa$_{\rm base}$} & +SimCSE & 70.16 & 81.77 & 73.24 & 81.36 & 80.65 & 80.22 & 68.56 & 76.57 \\
    & +DCLR & 70.01 & 83.08 & 75.09 & 83.66 & 81.06 & 81.86 & 70.33 & 77.87 \\
    & +DiffCSE & 70.05 & 83.43 & 75.49 & 82.81 & 82.12 & 82.38 & 71.19 & 78.21 \\
    & +PCL & 71.13 & 82.38 & 75.40 & 83.07 & 81.98 & 81.63 & 69.72 & 77.90 \\
    & \bf +\gmodelnospace$_{\rm listNet}$ & \underline{72.91} & \underline{85.72} & \underline{76.94} & \underline{84.52} &  \bf 82.59 & \bf 83.46 & \bf 71.94 & \underline{79.73} \\
    & \bf +\gmodelnospace$_{\rm listMLE}$ & \bf 73.20 & \bf 85.95 & \bf 77.17 & \bf 84.82 & \underline{82.58} & \underline{83.08} & \underline{71.88} & \bf 79.81 \\
    \midrule
    \multirow{5}{*}{RoBERTa$_{\rm large}$} & +SimCSE & 72.86 & 83.99 & 75.62 & 84.77 & 81.80 & 81.98 & 71.26 & 78.90 \\
    & +DCLR & 73.09 & 84.57 & 76.13 & 85.15 & 81.99 & 82.35 & 71.80 & 79.30 \\
    & +PCL & \bf 74.08 & 84.36 & 76.42 & 85.49 & 81.76 & 82.79 & 71.51 & 79.49 \\
    & \bf +\gmodelnospace$_{\rm listNet}$ & \underline{73.47} & \underline{85.77} & \bf 78.07 & \bf 85.65 & \underline{82.51} & \underline{84.12} & \bf 73.73 & \bf 80.47 \\
    & \bf +\gmodelnospace$_{\rm listMLE}$ & 73.20 & \bf 85.83 & \underline{78.00} & \underline{85.63} & \bf 82.67 & \bf 84.19 & \underline{73.64} &  \underline{80.45} \\
    \bottomrule
\end{tabular}
\end{adjustbox}
\caption{Sentence representations performance on STS tasks (Spearman's correlation). We directly import the results from the original papers and mark the best (bold) and second-best (underlined) results among models with the same PLMs. Results are statistically significant with respect to all baselines on each PLM (all p-value < 0.005).}
\label{tab:stsr}
\vspace{-3mm}
\end{table*}

\begin{table*}[t]
\small
\begin{adjustbox}{width=1.95\columnwidth,center}
\centering
  \begin{tabular}{p{2cm}|l|cccccccc}
    \toprule
    \bf PLMs & \bf Methods  & \bf MR & \bf CR & \bf SUBJ & \bf MPQA & \bf SST & \bf TREC & \bf MRPC & \bf avg. \\
    \midrule
    \multirow{2}{*}{Non-BERT} & GloVe(avg.) & 77.25 & 78.30 & 91.17 & 87.85 & 80.18 & 83.00 & 72.87 & 81.52 \\ 
    & Skip-thought & 76.50 & 80.10 & 93.60 & 87.10 & 82.00 & 92.20 & 73.00 & 83.50 \\
    \midrule
    \multirow{8}{*}{BERT$_{\rm base}$} & last avg. & 78.66 & 86.25 & 94.37 & 88.66 & 84.40 & \bf 92.80 & 69.54 & 84.94 \\
    & +IS & 81.09 & 87.18 & 94.96 & 88.75 & 85.96 & 88.64 & 74.24 & 85.83 \\
    & +SimCSE & 81.18 & 86.46 & 94.45 & 88.88 & 85.50 & 89.80 & 74.43 & 85.81 \\
    & +ArcCSE & 79.91 & 85.25 & \bf 99.58 & 89.21 & 84.90 & 89.20 & 74.78 & 86.12 \\
    & +DiffCSE$^\dagger$ & 81.76 & 86.20 & 94.76 & 89.21 & 86.00 & 87.60 & 75.54 & 85.87 \\
    & +PCL & 80.11 & 85.25 & 94.22 & 89.15 & 85.12 & 87.40 & 76.12 & 85.34 \\
    & {\bf +\gmodelnospace$_{\rm listNet}$} & \bf 83.21 & \underline{88.08} &  \underline{95.25} & \bf 90.00 & \bf 88.58 & \underline{90.00} & \underline{76.17} & \bf 87.33 \\
    & {\bf +\gmodelnospace$_{\rm listMLE}$} & \underline{83.07} & \bf 88.27 &  95.06 &  \underline{89.90} & \underline{87.70} &  89.40 &  \bf 76.23 &  \underline{87.09} \\
    \midrule
    \multirow{5}{*}{BERT$_{\rm large}$} & +SimCSE & \bf 85.36 & 89.38 & 95.39 & 89.63 & 90.44 & 91.80 & 76.41 & 88.34 \\
    & +ArcCSE & 84.34 & 88.82 & \bf 99.58 & 89.79 & 90.50 & 92.00 & 74.78 & 88.54 \\ 
    & +PCL & 82.47 & 87.87 & 95.04 & 89.59 & 87.75 & 93.00 & 76.00 & 87.39 \\
    & {\bf +\gmodelnospace$_{\rm listNet}$} & \underline{85.11} & \bf 89.56 & 95.39 & \bf 90.30 & \bf 90.77 & \bf 93.20 & \bf 77.16 & \bf 88.78 \\
    & {\bf +\gmodelnospace$_{\rm listMLE}$} & 84.63 & \underline{89.51} & \underline{95.50} & \underline{90.08} & \underline{90.61} & \bf 93.20 & \underline{76.99} & \underline{88.65} \\
    \midrule
    \multirow{5}{*}{RoBERTa$_{\rm base}$} & +SimCSE & 81.04 & 87.74 & 93.28 & 86.94 & 86.60 & 84.60 & 73.68 & 84.84 \\
    & +DiffCSE$^\dagger$ & 82.42 & 88.34 & 93.51 & 87.28 & 87.70 & 86.60 & 76.35 & 86.03 \\
    & +PCL & 81.83 & 87.55 & 92.92 & 87.21 & 87.26 & 85.20 & \underline{76.46} & 85.49 \\
    & {\bf +\gmodelnospace$_{\rm listNet}$} & \bf 83.53 & \bf 89.22 & \bf 94.07 & \bf 88.97 & \bf 89.95 & \underline{89.20} & \bf 76.52 & \bf 87.35 \\
    & {\bf +\gmodelnospace$_{\rm listMLE}$} & \underline{83.32} & \underline{88.61} & \underline{94.03} & \underline{88.88} & \underline{89.07} & \bf 90.80 & \underline{76.46} & \underline{87.31} \\
    \midrule
    \multirow{4}{*}{RoBERTa$_{\rm large}$} & +SimCSE & 82.74 & 87.87 & 93.66 & 88.22 & 88.58 & 92.00 & 69.68 & 86.11 \\
    & +PCL & \underline{84.47} & 89.06 & \underline{94.60} & 89.26 & 89.02 & \bf 94.20 & \underline{74.96} & \underline{87.94} \\
    & {\bf +\gmodelnospace$_{\rm listNet}$} & \underline{84.47} & \bf 89.51 & \bf 94.65 & \underline{89.87} & \underline{89.46} & \underline{93.00} & \bf 75.88 & \bf 88.12 \\
    & {\bf +\gmodelnospace$_{\rm listMLE}$} & \bf 84.61 & \underline{89.27} & 94.47 & \bf 89.99 & \bf 89.73 & 92.60 & 74.43 & 87.87 \\
    \bottomrule
\end{tabular}
\end{adjustbox}
\caption{Sentence representations performance on transfer tasks (accuracy). The results of DiffCSE$^\dagger$ are obtained from the publicly available code and checkpoints, while others are imported from the original papers. We mark the best (bold) and second-best (underlined) results among models with the same PLMs. Results are statistically significant with respect to all baselines on each PLM (all p-value < 0.005).}
\label{tab:tranr}
\vspace{-3mm}
\end{table*}

\subsection{Setup} \label{sec:setup}
We evaluate our approach on two sentence related tasks, Semantic Textual Similarity (STS) and Transfer (TR). The SentEval toolkit \cite{DBLP:conf/lrec/ConneauK18} is used in our experiments. For STS tasks, we evaluate on seven datasets: STS12-16 \cite{DBLP:conf/semeval/AgirreCDG12,DBLP:conf/starsem/AgirreCDGG13,DBLP:conf/semeval/AgirreBCCDGGMRW14,DBLP:conf/semeval/AgirreBCCDGGLMM15,DBLP:conf/semeval/AgirreBCDGMRW16}, STS Benchmark \cite{DBLP:conf/semeval/CerDALS17} and SICK-Relatedness \cite{DBLP:conf/lrec/MarelliMBBBZ14}. These datasets contain pairs of sentences with similarity score labels from 0 to 5. Following SimCSE, we directly compute the cosine similarity between the sentence representations which means all the STS experiments are fully unsupervised, and report the Spearman's correlation. For TR tasks, we evaluate on seven datasets with the default configurations from SentEval: MR \cite{DBLP:conf/acl/PangL05}, CR \cite{DBLP:conf/kdd/HuL04}, SUBJ \cite{DBLP:conf/acl/PangL04}, MPQA \cite{DBLP:journals/lre/WiebeWC05}, SST-2 \cite{DBLP:conf/emnlp/SocherPWCMNP13}, TREC \cite{DBLP:conf/sigir/VoorheesT00} and MRPC \cite{DBLP:conf/acl-iwp/DolanB05}. We use a logistic regression classifier trained on top of the frozen sentence representations, and report the classification accuracy.

For fair comparison, we use the same $10^6$ randomly sampled sentences from English Wikipedia provided by SimCSE. Following previous works, we start from pre-trained checkpoints of BERT \cite{DBLP:conf/naacl/DevlinCLT19} and RoBERTa \cite{DBLP:journals/corr/abs-1907-11692}, and utilize the embedding corresponding to [CLS] token as the representation of the input sentence. First we train SimCSE models including four variants: SimCSE-BERT$_{\rm base}$, SimCSE-BERT$_{\rm large}$, SimCSE-RoBERTa$_{\rm base}$ and SimCSE-RoBERTa$_{\rm large}$. We utilize the first two as a multi-teacher for \gmodelnospace-BERT$_{\rm base}$ and \gmodelnospace-BERT$_{\rm large}$, while the last two for \gmodelnospace-RoBERTa$_{\rm base}$ and \gmodelnospace-RoBERTa$_{\rm large}$. We evaluate our model every 125 training steps on the dev set of STS-B and keep the best checkpoint for the evaluation on test sets of all STS and TR tasks. More training details can be found in Appendix \ref{ap:detail}.

We compare \gmodel with several unsupervised sentence representation learning methods, including average GloVe embeddings \citep{DBLP:conf/emnlp/PenningtonSM14}, USE \citep{DBLP:conf/emnlp/CerYKHLJCGYTSK18} and Skip-thought \citep{DBLP:conf/nips/KirosZSZUTF15}, average BERT embeddings from the last layer, post-processing methods such as BERT-flow \citep{DBLP:conf/emnlp/LiZHWYL20} and BERT-whitening \citep{DBLP:journals/corr/abs-2103-15316}, and contrastive learning methods such as IS-BERT \citep{DBLP:conf/emnlp/ZhangHLLB20} and ConSERT \citep{DBLP:conf/acl/YanLWZWX20}. We also include recent strong unsupervised sentence representation baselines, including SimCSE \cite{DBLP:conf/emnlp/GaoYC21}, DCLR \cite{DBLP:conf/acl/ZhouZZW22}, ArcCSE \cite{DBLP:conf/acl/ZhangZWXLZ22}, DiffCSE \cite{DBLP:conf/naacl/ChuangDLZCS0YKG22}, PaSER \cite{DBLP:conf/emnlp/WuZ22} and PCL \cite{DBLP:journals/corr/abs-2201-12093}. Since \gmodel and the teacher model SimCSE are using the same unsupervised training data, the comparison between \gmodel and baselines is fair.

\subsection{Main Results}
\paragraph{Results on STS Tasks} As shown in Table \ref{tab:stsr}, it is clear that \gmodel significantly outperforms the previous methods on all PLMs, which demonstrates the effectiveness of our approach. For example, compared with SimCSE, \gmodel has brought noticeable improvements: 4.11\% on BERT$_{\rm base}$, 2.19\% on BERT$_{\rm large}$, 3.24\% on RoBERTa$_{\rm base}$ and 1.57\% on RoBERTa$_{\rm large}$. \gmodelnospace-BERT$_{\rm base}$ even outperforms SimCSE-BERT$_{\rm large}$ by nearly 2\%. Compared with the previous state-of-the-art methods, \gmodel still achieves consistent improvements, which validates that \gmodel is able to obtain more semantically discriminative representations by incorporating ranking consistency and ranking distillation. We also observe that the performances of \gmodelnospace$_{\rm listNet}$ and \gmodelnospace$_{\rm listMLE}$ are very consistent across all datasets, which demonstrates the effectiveness of both listwise ranking methods.
\begin{table}[t]
\small
\tabcolsep=0.1cm
\centering
\begin{adjustbox}{width=0.98\columnwidth,center}
  \begin{tabular}{lcc}
    \toprule
    \bf Models & \bf STS(avg.) & \bf TR(avg.) \\
    \midrule
    SimCSE & 76.25 & 85.81 \\
    \midrule
    \gmodelnospace$_{\rm listNet}$ & 80.05 & 87.33 \\
    \enspace w/o $\mathcal{L}_{\rm{consistency}}$ & 79.56 & 86.80\\
    \enspace w/o $\mathcal{L}_{\rm{infoNCE}}$ & 79.72 & 86.91 \\
    \enspace w/o $\mathcal{L}_{\rm{consistency}}$,$\mathcal{L}_{\rm{infoNCE}}$ & 79.41 & 86.76 \\
    \midrule
    \gmodelnospace$_{\rm listMLE}$ & 80.36 & 87.09 \\
    \enspace w/o $\mathcal{L}_{\rm{consistency}}$ & 79.88 & 86.65 \\
    \enspace w/o $\mathcal{L}_{\rm{infoNCE}}$ & 79.95 & 86.73 \\
    \enspace w/o $\mathcal{L}_{\rm{consistency}}$,$\mathcal{L}_{\rm{infoNCE}}$ & 79.73 & 86.24 \\
    \midrule
    \gmodel w/o  $\mathcal{L}_{\rm{rank}}$ & 76.93 & 85.97\\
    \gmodel w/o  $\mathcal{L}_{\rm{infoNCE}}$, $\mathcal{L}_{\rm{rank}}$ & 73.74 & 85.56\\
    \bottomrule
\end{tabular}
\end{adjustbox}
\caption{Ablation studies of different loss functions based on BERT$_{\rm base}$. Other PLMs yield similar patterns to BERT$_{\rm base}$.}
\label{tab:ablation}
\vspace{-3mm}
\end{table}

\paragraph{Results on TR Tasks} 
It can be seen in Table \ref{tab:tranr} that \gmodel achieves the best performance among all the compared baselines on all PLMs. Note that for DiffCSE, we obtain the results from the publicly available code and checkpoints, because DiffCSE uses different dev sets to find the best hyperparameters for TR tasks than other baselines. More detailed explanation and comprehensive comparison are provided in Appendix \ref{ap:tr}. Another observation is that the performance of the \gmodelnospace$_{\rm listNet}$ is slightly better than that of the \gmodelnospace$_{\rm listMLE}$. Our hypothesis is that the inaccurate pseudo ranking labels introduce more errors in the calculation of the permutation probability than the top one probability. Nevertheless, both listwise methods achieve better results than the baselines, which is consistent with the results in Table \ref{tab:stsr}.
\begin{table}[t]
\small
\centering
\begin{adjustbox}{width=0.98\columnwidth,center}
  \begin{tabular}{c|cc}
    \toprule
    \multirow{2}{*}{\bf Teacher} &  \multicolumn{2}{c}{\bf \gmodelnospace} \\ 
    & ListNet & ListMLE \\
    \midrule
    SimCSE$_{\rm base}$ & 77.48 & 77.75 \\ 
    DiffCSE$_{\rm base}$ & 78.87 & 79.06 \\ 
    SimCSE$_{\rm large}$ & 79.66 & 79.81 \\ 
    SimCSE$_{\rm base}$+DiffCSE$_{\rm base}$ & 79.10 & 79.28 \\ 
    SimCSE$_{\rm base}$+SimCSE$_{\rm large}$ & 80.05 & 80.36 \\ 
    DiffCSE$_{\rm base}$+SimCSE$_{\rm large}$ & 80.20 & 80.47 \\
    \bottomrule
\end{tabular}
\end{adjustbox}
\caption{Comparisons of different teachers based on BERT. Results of \gmodel are average STS performance using BERT$_{\rm base}$.}
\label{tab:teacher}
\vspace{-3mm}
\end{table}

\begin{figure*}[t]
    \centering
	\begin{minipage}{0.56\linewidth}
	    \centering
		\subfloat[\gmodelnospace$_{\rm listNet}$]{\includegraphics[width=0.495\textwidth]{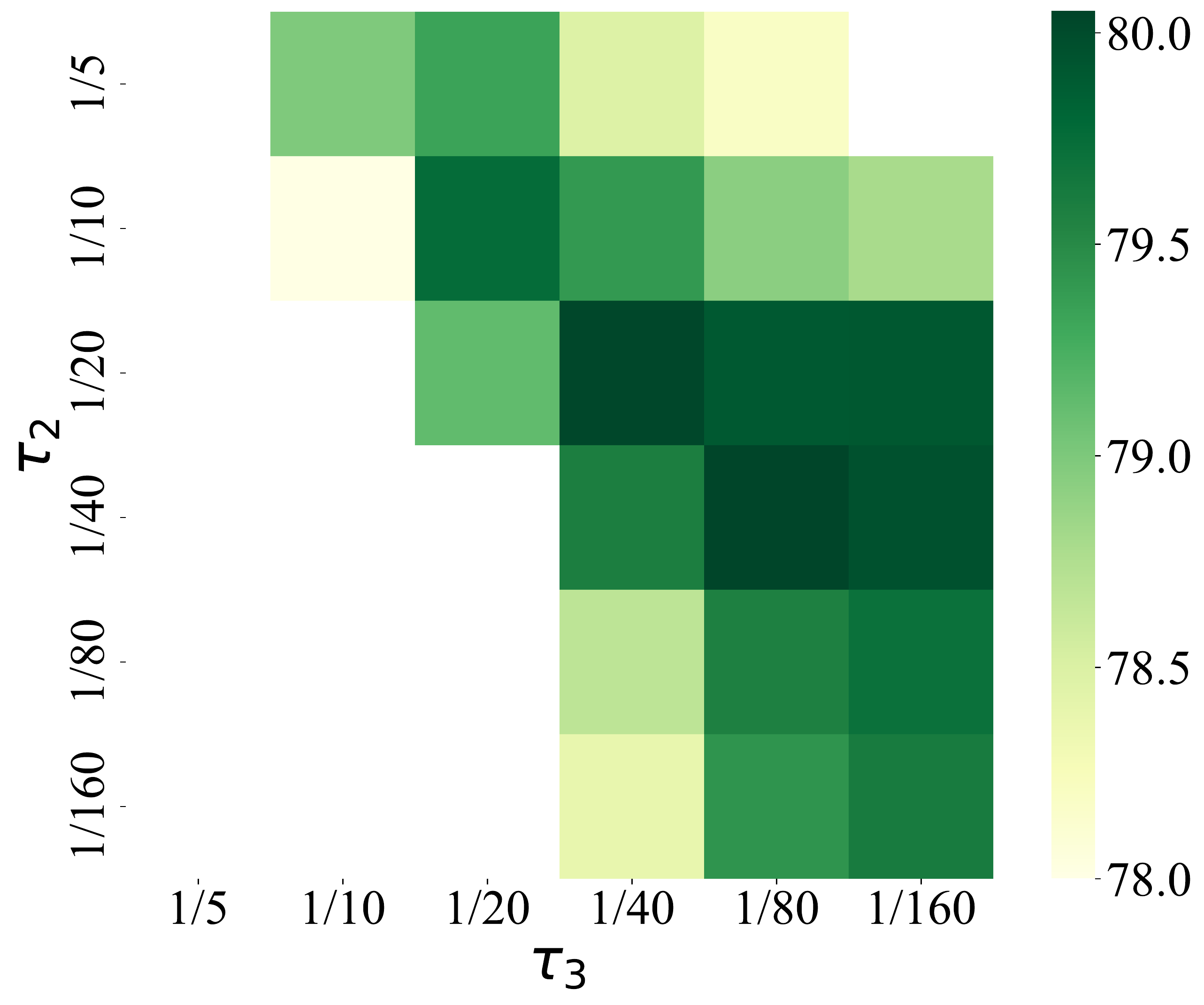}\label{fig:listnet_temp}}
\subfloat[\gmodelnospace$_{\rm listMLE}$]{\includegraphics[width=0.495\textwidth]{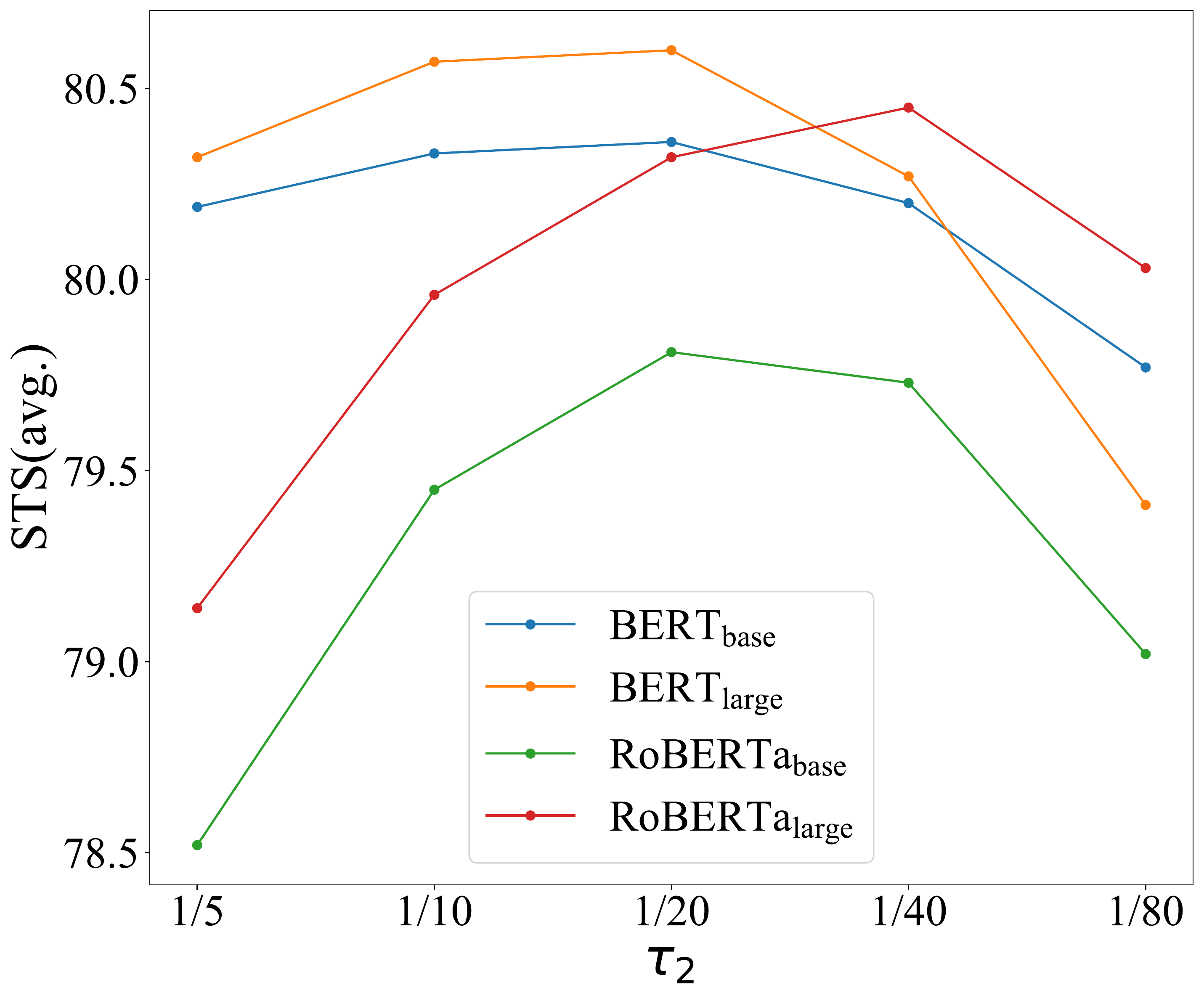}\label{fig:listmle_temp}}
		\caption{Effect of the temperatures $\tau_2$ and $\tau_3$. Results are average STS performance, and \gmodelnospace$_{\rm listNet}$ is based on BERT$_{\rm base}$ while \gmodelnospace$_{\rm listMLE}$ is based on different PLMs. We do not demonstrate results below 78 to make the variation obvious.}
        \label{fig:temp}
        \vspace{-2.5mm}
	\end{minipage}
	\hspace{0.1cm}
	\begin{minipage}{0.42\linewidth}
	    \centering
		\includegraphics[width=1.0\columnwidth]{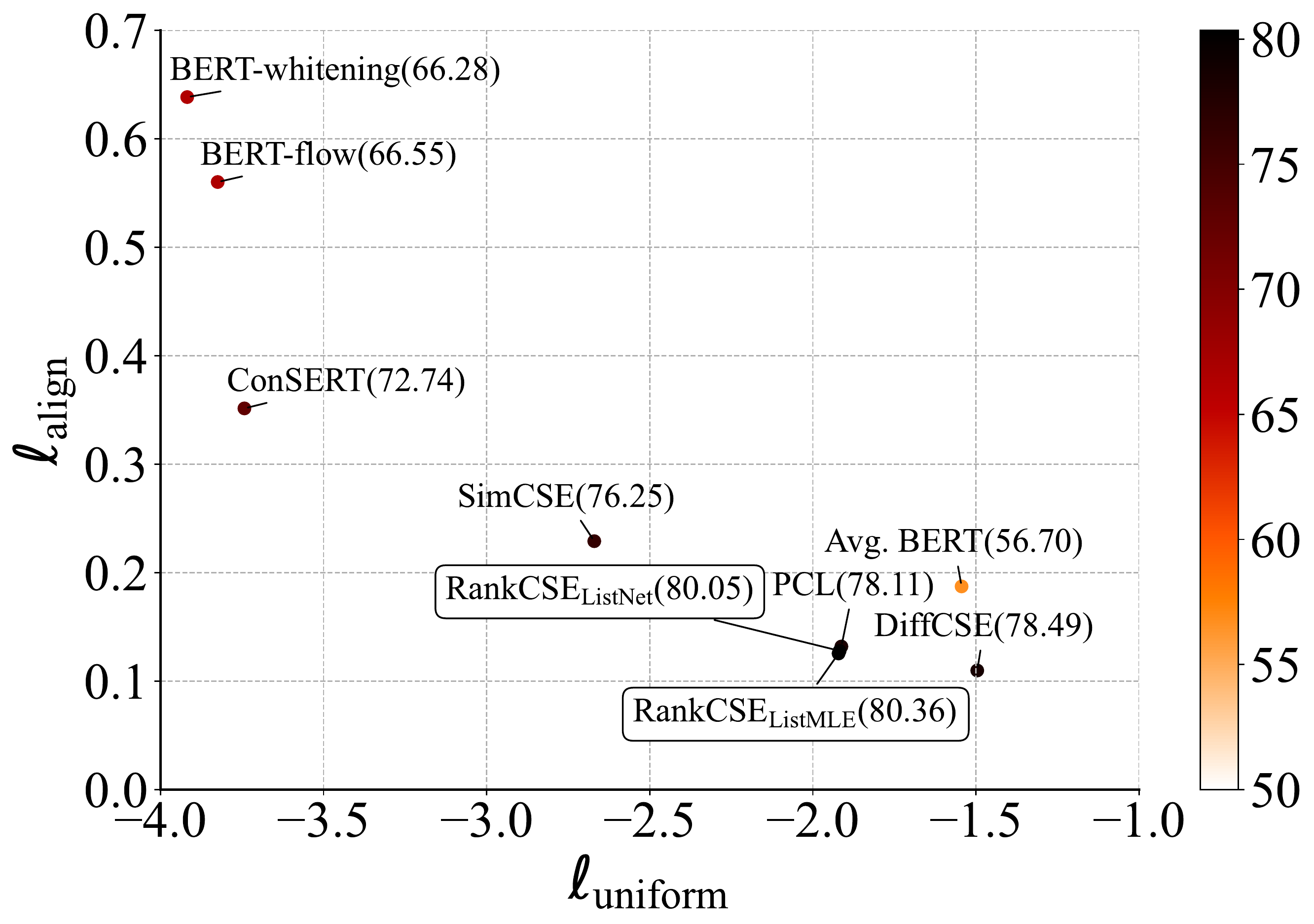}
        \caption{
    $\ell_{\rm align}$-$\ell_{\rm uniform}$ plot for different sentence
representation methods based on BERT$_{\rm base}$ measured on the STS-B dev set. Color of points represents average STS performance.}
    \label{fig:align-uniform}
	\end{minipage}
\end{figure*}
\begin{table*}[t]
\begin{adjustbox}{width=1.95\columnwidth,center}
\centering
  \begin{tabular}{c|cccccc}
    \toprule
    \multirow{2}{*}{\bf PLMs} & \multicolumn{2}{c}{\bf \gmodelnospace$_{\rm listNet}$} & \multicolumn{2}{c}{\bf \gmodelnospace$_{\rm listMLE}$} & \multicolumn{2}{c}{\bf SimCSE}\\
    & STS(avg.) & TR(avg.) & STS(avg.) & TR(avg.) & STS(avg.) & TR(avg.) \\
    \midrule
    BERT$_{\rm base}$ & 80.00$\pm$0.13& 87.28$\pm$0.19& 80.39$\pm$0.04 & 87.05$\pm$0.06 & 75.52$\pm$0.70 & 85.44$\pm$0.47 \\
    BERT$_{\rm large}$ & 80.41$\pm$0.10& 88.74$\pm$0.14& 80.59$\pm$0.05 & 88.63$\pm$0.06 & 77.79$\pm$0.64 & 88.10$\pm$0.36 \\
    RoBERTa$_{\rm base}$ & 79.67$\pm$0.09& 87.46$\pm$0.13& 79.78$\pm$0.05& 87.30$\pm$0.07 & 76.45$\pm$0.56 & 84.74$\pm$0.38 \\
    RoBERTa$_{\rm large}$ & 80.46$\pm$0.11 & 87.97$\pm$0.14 & 80.34$\pm$0.08 & 87.82$\pm$0.08 & 78.53$\pm$0.49 & 86.29$\pm$0.33\\
    \bottomrule
\end{tabular}
\end{adjustbox}
\caption{Mean and standard deviation across five different runs of \gmodel and SimCSE.}
\label{tab:mean_variance}
\vspace{-3mm}
\end{table*}
\subsection{Analysis and Discussion} \label{sec5.3}
\paragraph{Ablation Study} To investigate the impact of different losses in our approach, we conduct a set of ablation studies by removing $\mathcal{L_{\rm infoNCE}}$, $\mathcal{L}_{\rm consistency}$ and $\mathcal{L}_{\rm rank}$ from Eq.(\ref{eq:totloss}). The average results on STS and TR tasks are reported in Table \ref{tab:ablation}. There are several observations from the results. First, when $\mathcal{L}_{\rm rank}$ is removed, the performance significantly drops in both STS and TR tasks, which indicates the effectiveness of $\mathcal{L}_{\rm rank}$ in our modeling. Second, it is also clear that without $\mathcal{L}_{\rm infoNCE}$ or $\mathcal{L}_{\rm consistency}$, the model performance also decreases, especially on TR tasks. Thirdly, it is worth mentioning that \gmodel with only $\mathcal{L}_{\rm rank}$ can also outperform the teachers on STS tasks. The reason is that RankCSE is able to preserve ranking knowledge from multiple teachers, and generalize fine-grained ranking information from multiple coarse-grained representations. Fourthly, since $\mathcal{L}_{\rm consistency}$ does not explicitly distinguish the positives from negatives, \gmodel with only $\mathcal{L}_{\rm consistency}$ will preserve inaccurate rankings leading to significant performance drop. Finally, \gmodel with all components achieves the best performance on both STS and TR tasks.

\paragraph{Comparisons of Different Teachers} We conduct experiments to explore the impact of different teachers on the performance of \gmodelnospace. As shown in Table \ref{tab:teacher}, \gmodel outperforms the teacher model which indicates that incorporating ranking consistency and ranking distillation leads to more semantically discriminative sentence representations. Comparing the performance of \gmodel using different teachers, we observe that better teacher leads to better \gmodelnospace, which is consistent with our expectation since accurate ranking labels yield more effective ranking knowledge transfer. Another observation is that the performance of \gmodel with a multi-teacher is better than that with a single teacher, which verifies that \gmodel is able to preserve listwise ranking knowledge from more than one teacher. It is also interesting to see that using DiffCSE-BERT$_{\rm base}$ and SimCSE-BERT$_{\rm large}$ as multi-teacher leads to even higher performance than the results in Table \ref{tab:stsr}. We plan to conduct more investigation along this direction to explore the upper bound of improvements.

\paragraph{Effect of Hyperparameters} To study the effect of temperature hyperparameters, we conduct experiments by setting different $\tau_2$ and $\tau_3$. As shown in Figure \ref{fig:listnet_temp}, we find that large discrepancy between $\tau_2$ and $\tau_3$ leads to significant drop in the performance of \gmodelnospace$_{\rm ListNet}$. The best temperature setting for \gmodelnospace$_{\rm ListNet}$ is $\tau_2 : \tau_3 = 2 : 1$. The performance of \gmodelnospace$_{\rm ListMLE}$ has similar trends based on different PLMs, as shown in Figure \ref{fig:listmle_temp}. For both \gmodelnospace$_{\rm ListNet}$ and \gmodelnospace$_{\rm ListMLE}$, the temperature should be set moderate.

\paragraph{Robustness of \gmodelnospace} We conduct 5 runs of model training with the hyperparameter settings which can be referred to Appendix \ref{ap:detail} with different random seeds, and then calculate the mean and standard deviation values. The results provided in Table \ref{tab:mean_variance} demonstrate both the superior performance and the robustness of our model. It can also be seen that \gmodelnospace$_{\rm listMLE}$ achieves similar performance but more stable results compared with \gmodelnospace$_{\rm listNet}$.

\paragraph{Alignment and Uniformity} Following previous works \cite{DBLP:conf/icml/0001I20}, we use alignment and uniformity to measure the quality of representation space.
Alignment measures the distance between similar instances, while uniformity measures how well the representations are uniformly distributed (detailed in Appendix \ref{ap:aligment-uniformity}). For both measures, the smaller value indicates the better result. We plot the distribution of $\ell_{\rm align}$-$\ell_{\rm uniform}$ for different models using BERT$_{\rm base}$ which are measured on the STS-B dev set. As shown in Figure \ref{fig:align-uniform}, \gmodel effectively improves both alignment and uniformity compared with average BERT embeddings, while SimCSE and DiffCSE only improve uniformity and alignment respectively. Since \gmodel pulls similar negatives closer during incorporating ranking consistency and ranking distillation, \gmodel has smaller alignment and bigger uniformity than SimCSE. We consider that \gmodel achieves a better trade-off than SimCSE. When compared with DiffCSE, \gmodel has smaller uniformity whereas similar alignment. We can also observe that \gmodel outperforms PCL on both metrics.

\section{Conclusion}
In this work, we propose \gmodelnospace, an unsupervised approach to learn more semantically discriminative sentence representations. The core idea of \gmodel is incorporating ranking consistency and ranking distillation with contrastive learning into a unified framework. When simultaneously ensuring ranking consistency and distilling listwise ranking knowledge from the teacher, \gmodel can learn how to make fine-grained distinctions in semantics, leading to more semantically discriminative sentence representations. Experimental results on STS and TR tasks demonstrate that \gmodel outperforms previous state-of-the-art methods. We also conduct thorough ablation study and analysis to demonstrate the effectiveness of each component and justify the inner workings of our approach. We leave what is the upper bound of improvements of the teacher for future work.

\section*{Limitations}
In this section, we discuss the limitations of our work as follows. First, despite achieving promising results, our model needs to calculate pseudo ranking labels of the teacher which requires additional training time per epoch than the teacher. The training efficiency of \gmodel and SimCSE can be seen in Appendix \ref{ap:efficiency}. Second, we directly use SimCSE$_{\rm base}$ and SimCSE$_{\rm large}$ as a multi-teacher in our implementation and experiments. However, how to choose the best combination of the teacher models is worth further exploration. It could help researchers to better understand the upper bound of improvements. We plan to investigate more along this direction in the future.

\section*{Acknowledgements}
This work is supported by Ministry of Science and Technology Key R\&D Program (2030 Artificial Intelligence) (No. 2020AAA0106600) and National Natural Science Foundation of China (NSFC Grant No. 62122089). We sincerely thank all reviewers for their valuable comments and suggestions, which are crucial for improving our work. We would also like to acknowledge Angela Li for her contributions in creating the figures used in this work.

\bibliography{anthology,custom}

\begin{thebibliography}{55}
\expandafter\ifx\csname natexlab\endcsname\relax\def\natexlab#1{#1}\fi

\bibitem[{Abdi(2007)}]{abdi2007kendall}
Herv{\'e} Abdi. 2007.
\newblock \href
  {https://personal.utdallas.edu/~Herve/Abdi-KendallCorrelation2007-pretty.pdf}
  {The kendall rank correlation coefficient}.
\newblock \emph{Encyclopedia of measurement and statistics}, 2:508--510.

\bibitem[{Agirre et~al.(2015)Agirre, Banea, Cardie, Cer, Diab,
  Gonzalez{-}Agirre, Guo, Lopez{-}Gazpio, Maritxalar, Mihalcea, Rigau, Uria,
  and Wiebe}]{DBLP:conf/semeval/AgirreBCCDGGLMM15}
Eneko Agirre, Carmen Banea, Claire Cardie, Daniel~M. Cer, Mona~T. Diab, Aitor
  Gonzalez{-}Agirre, Weiwei Guo, I{\~{n}}igo Lopez{-}Gazpio, Montse Maritxalar,
  Rada Mihalcea, German Rigau, Larraitz Uria, and Janyce Wiebe. 2015.
\newblock \href {https://doi.org/10.18653/v1/s15-2045} {Semeval-2015 task 2:
  Semantic textual similarity, english, spanish and pilot on interpretability}.
\newblock In \emph{Proceedings of the 9th International Workshop on Semantic
  Evaluation, SemEval@NAACL-HLT 2015, Denver, Colorado, USA, June 4-5, 2015},
  pages 252--263. The Association for Computer Linguistics.

\bibitem[{Agirre et~al.(2014)Agirre, Banea, Cardie, Cer, Diab,
  Gonzalez{-}Agirre, Guo, Mihalcea, Rigau, and
  Wiebe}]{DBLP:conf/semeval/AgirreBCCDGGMRW14}
Eneko Agirre, Carmen Banea, Claire Cardie, Daniel~M. Cer, Mona~T. Diab, Aitor
  Gonzalez{-}Agirre, Weiwei Guo, Rada Mihalcea, German Rigau, and Janyce Wiebe.
  2014.
\newblock \href {https://doi.org/10.3115/v1/s14-2010} {Semeval-2014 task 10:
  Multilingual semantic textual similarity}.
\newblock In \emph{Proceedings of the 8th International Workshop on Semantic
  Evaluation, SemEval@COLING 2014, Dublin, Ireland, August 23-24, 2014}, pages
  81--91. The Association for Computer Linguistics.

\bibitem[{Agirre et~al.(2016)Agirre, Banea, Cer, Diab, Gonzalez{-}Agirre,
  Mihalcea, Rigau, and Wiebe}]{DBLP:conf/semeval/AgirreBCDGMRW16}
Eneko Agirre, Carmen Banea, Daniel~M. Cer, Mona~T. Diab, Aitor
  Gonzalez{-}Agirre, Rada Mihalcea, German Rigau, and Janyce Wiebe. 2016.
\newblock \href {https://doi.org/10.18653/v1/s16-1081} {Semeval-2016 task 1:
  Semantic textual similarity, monolingual and cross-lingual evaluation}.
\newblock In \emph{Proceedings of the 10th International Workshop on Semantic
  Evaluation, SemEval@NAACL-HLT 2016, San Diego, CA, USA, June 16-17, 2016},
  pages 497--511. The Association for Computer Linguistics.

\bibitem[{Agirre et~al.(2012)Agirre, Cer, Diab, and
  Gonzalez{-}Agirre}]{DBLP:conf/semeval/AgirreCDG12}
Eneko Agirre, Daniel~M. Cer, Mona~T. Diab, and Aitor Gonzalez{-}Agirre. 2012.
\newblock \href {https://aclanthology.org/S12-1051/} {Semeval-2012 task 6: {A}
  pilot on semantic textual similarity}.
\newblock In \emph{Proceedings of the 6th International Workshop on Semantic
  Evaluation, SemEval@NAACL-HLT 2012, Montr{\'{e}}al, Canada, June 7-8, 2012},
  pages 385--393. The Association for Computer Linguistics.

\bibitem[{Agirre et~al.(2013)Agirre, Cer, Diab, Gonzalez{-}Agirre, and
  Guo}]{DBLP:conf/starsem/AgirreCDGG13}
Eneko Agirre, Daniel~M. Cer, Mona~T. Diab, Aitor Gonzalez{-}Agirre, and Weiwei
  Guo. 2013.
\newblock \href {https://aclanthology.org/S13-1004/} {*sem 2013 shared task:
  Semantic textual similarity}.
\newblock In \emph{Proceedings of the Second Joint Conference on Lexical and
  Computational Semantics, *SEM 2013, June 13-14, 2013, Atlanta, Georgia,
  {USA}}, pages 32--43. Association for Computational Linguistics.

\bibitem[{Burges et~al.(2006)Burges, Ragno, and Le}]{DBLP:conf/nips/BurgesRL06}
Christopher J.~C. Burges, Robert Ragno, and Quoc~Viet Le. 2006.
\newblock \href
  {https://proceedings.neurips.cc/paper/2006/hash/af44c4c56f385c43f2529f9b1b018f6a-Abstract.html}
  {Learning to rank with nonsmooth cost functions}.
\newblock In \emph{Advances in Neural Information Processing Systems 19,
  Proceedings of the Twentieth Annual Conference on Neural Information
  Processing Systems, Vancouver, British Columbia, Canada, December 4-7, 2006},
  pages 193--200. {MIT} Press.

\bibitem[{Burges et~al.(2005)Burges, Shaked, Renshaw, Lazier, Deeds, Hamilton,
  and Hullender}]{DBLP:conf/icml/BurgesSRLDHH05}
Christopher J.~C. Burges, Tal Shaked, Erin Renshaw, Ari Lazier, Matt Deeds,
  Nicole Hamilton, and Gregory~N. Hullender. 2005.
\newblock \href {https://doi.org/10.1145/1102351.1102363} {Learning to rank
  using gradient descent}.
\newblock In \emph{Machine Learning, Proceedings of the Twenty-Second
  International Conference {(ICML} 2005), Bonn, Germany, August 7-11, 2005},
  volume 119 of \emph{{ACM} International Conference Proceeding Series}, pages
  89--96. {ACM}.

\bibitem[{Cao et~al.(2007)Cao, Qin, Liu, Tsai, and
  Li}]{DBLP:conf/icml/CaoQLTL07}
Zhe Cao, Tao Qin, Tie{-}Yan Liu, Ming{-}Feng Tsai, and Hang Li. 2007.
\newblock \href {https://doi.org/10.1145/1273496.1273513} {Learning to rank:
  from pairwise approach to listwise approach}.
\newblock In \emph{Machine Learning, Proceedings of the Twenty-Fourth
  International Conference {(ICML} 2007), Corvallis, Oregon, USA, June 20-24,
  2007}, volume 227 of \emph{{ACM} International Conference Proceeding Series},
  pages 129--136. {ACM}.

\bibitem[{Carlsson et~al.(2021)Carlsson, Gyllensten, Gogoulou, Hellqvist, and
  Sahlgren}]{DBLP:conf/iclr/CarlssonGGHS21}
Fredrik Carlsson, Amaru~Cuba Gyllensten, Evangelia Gogoulou,
  Erik~Ylip{\"{a}}{\"{a}} Hellqvist, and Magnus Sahlgren. 2021.
\newblock \href {https://openreview.net/forum?id=Ov\_sMNau-PF} {Semantic
  re-tuning with contrastive tension}.
\newblock In \emph{9th International Conference on Learning Representations,
  {ICLR} 2021, Virtual Event, Austria, May 3-7, 2021}. OpenReview.net.

\bibitem[{Cer et~al.(2018)Cer, Yang, Kong, Hua, Limtiaco, John, Constant,
  Guajardo{-}Cespedes, Yuan, Tar, Strope, and
  Kurzweil}]{DBLP:conf/emnlp/CerYKHLJCGYTSK18}
Daniel Cer, Yinfei Yang, Sheng{-}yi Kong, Nan Hua, Nicole Limtiaco, Rhomni~St.
  John, Noah Constant, Mario Guajardo{-}Cespedes, Steve Yuan, Chris Tar, Brian
  Strope, and Ray Kurzweil. 2018.
\newblock \href {https://doi.org/10.18653/v1/d18-2029} {Universal sentence
  encoder for english}.
\newblock In \emph{Proceedings of the 2018 Conference on Empirical Methods in
  Natural Language Processing, {EMNLP} 2018: System Demonstrations, Brussels,
  Belgium, October 31 - November 4, 2018}, pages 169--174. Association for
  Computational Linguistics.

\bibitem[{Cer et~al.(2017)Cer, Diab, Agirre, Lopez{-}Gazpio, and
  Specia}]{DBLP:conf/semeval/CerDALS17}
Daniel~M. Cer, Mona~T. Diab, Eneko Agirre, I{\~{n}}igo Lopez{-}Gazpio, and
  Lucia Specia. 2017.
\newblock \href {https://doi.org/10.18653/v1/S17-2001} {Semeval-2017 task 1:
  Semantic textual similarity multilingual and crosslingual focused
  evaluation}.
\newblock In \emph{Proceedings of the 11th International Workshop on Semantic
  Evaluation, SemEval@ACL 2017, Vancouver, Canada, August 3-4, 2017}, pages
  1--14. Association for Computational Linguistics.

\bibitem[{Chen et~al.(2017)Chen, Sun, Shi, and Hong}]{DBLP:conf/kdd/ChenSSH17}
Ting Chen, Yizhou Sun, Yue Shi, and Liangjie Hong. 2017.
\newblock \href {https://doi.org/10.1145/3097983.3098202} {On sampling
  strategies for neural network-based collaborative filtering}.
\newblock In \emph{Proceedings of the 23rd {ACM} {SIGKDD} International
  Conference on Knowledge Discovery and Data Mining, Halifax, NS, Canada,
  August 13 - 17, 2017}, pages 767--776. {ACM}.

\bibitem[{Chuang et~al.(2022)Chuang, Dangovski, Luo, Zhang, Chang, Soljacic,
  Li, Yih, Kim, and Glass}]{DBLP:conf/naacl/ChuangDLZCS0YKG22}
Yung{-}Sung Chuang, Rumen Dangovski, Hongyin Luo, Yang Zhang, Shiyu Chang,
  Marin Soljacic, Shang{-}Wen Li, Scott Yih, Yoon Kim, and James~R. Glass.
  2022.
\newblock \href {https://doi.org/10.18653/v1/2022.naacl-main.311} {Diffcse:
  Difference-based contrastive learning for sentence embeddings}.
\newblock In \emph{Proceedings of the 2022 Conference of the North American
  Chapter of the Association for Computational Linguistics: Human Language
  Technologies, {NAACL} 2022, Seattle, WA, United States, July 10-15, 2022},
  pages 4207--4218. Association for Computational Linguistics.

\bibitem[{Conneau and Kiela(2018)}]{DBLP:conf/lrec/ConneauK18}
Alexis Conneau and Douwe Kiela. 2018.
\newblock \href
  {http://www.lrec-conf.org/proceedings/lrec2018/summaries/757.html} {Senteval:
  An evaluation toolkit for universal sentence representations}.
\newblock In \emph{Proceedings of the Eleventh International Conference on
  Language Resources and Evaluation, {LREC} 2018, Miyazaki, Japan, May 7-12,
  2018}. European Language Resources Association {(ELRA)}.

\bibitem[{Devlin et~al.(2019)Devlin, Chang, Lee, and
  Toutanova}]{DBLP:conf/naacl/DevlinCLT19}
Jacob Devlin, Ming{-}Wei Chang, Kenton Lee, and Kristina Toutanova. 2019.
\newblock \href {https://doi.org/10.18653/v1/n19-1423} {{BERT:} pre-training of
  deep bidirectional transformers for language understanding}.
\newblock In \emph{Proceedings of the 2019 Conference of the North American
  Chapter of the Association for Computational Linguistics: Human Language
  Technologies, {NAACL-HLT} 2019, Minneapolis, MN, USA, June 2-7, 2019, Volume
  1 (Long and Short Papers)}, pages 4171--4186. Association for Computational
  Linguistics.

\bibitem[{Dolan and Brockett(2005)}]{DBLP:conf/acl-iwp/DolanB05}
William~B. Dolan and Chris Brockett. 2005.
\newblock \href {https://aclanthology.org/I05-5002/} {Automatically
  constructing a corpus of sentential paraphrases}.
\newblock In \emph{Proceedings of the Third International Workshop on
  Paraphrasing, IWP@IJCNLP 2005, Jeju Island, Korea, October 2005, 2005}. Asian
  Federation of Natural Language Processing.

\bibitem[{Ethayarajh(2019)}]{DBLP:conf/emnlp/Ethayarajh19}
Kawin Ethayarajh. 2019.
\newblock \href {https://doi.org/10.18653/v1/D19-1006} {How contextual are
  contextualized word representations? comparing the geometry of bert, elmo,
  and {GPT-2} embeddings}.
\newblock In \emph{Proceedings of the 2019 Conference on Empirical Methods in
  Natural Language Processing and the 9th International Joint Conference on
  Natural Language Processing, {EMNLP-IJCNLP} 2019, Hong Kong, China, November
  3-7, 2019}, pages 55--65. Association for Computational Linguistics.

\bibitem[{Gao et~al.(2021)Gao, Yao, and Chen}]{DBLP:conf/emnlp/GaoYC21}
Tianyu Gao, Xingcheng Yao, and Danqi Chen. 2021.
\newblock \href {https://doi.org/10.18653/v1/2021.emnlp-main.552} {Simcse:
  Simple contrastive learning of sentence embeddings}.
\newblock In \emph{Proceedings of the 2021 Conference on Empirical Methods in
  Natural Language Processing, {EMNLP} 2021, Virtual Event / Punta Cana,
  Dominican Republic, 7-11 November, 2021}, pages 6894--6910. Association for
  Computational Linguistics.

\bibitem[{Giorgi et~al.(2021)Giorgi, Nitski, Wang, and
  Bader}]{DBLP:conf/acl/GiorgiNWB20}
John~M. Giorgi, Osvald Nitski, Bo~Wang, and Gary~D. Bader. 2021.
\newblock \href {https://doi.org/10.18653/v1/2021.acl-long.72} {Declutr: Deep
  contrastive learning for unsupervised textual representations}.
\newblock In \emph{Proceedings of the 59th Annual Meeting of the Association
  for Computational Linguistics and the 11th International Joint Conference on
  Natural Language Processing, {ACL/IJCNLP} 2021, (Volume 1: Long Papers),
  Virtual Event, August 1-6, 2021}, pages 879--895. Association for
  Computational Linguistics.

\bibitem[{Hill et~al.(2016)Hill, Cho, and Korhonen}]{DBLP:conf/naacl/HillCK16}
Felix Hill, Kyunghyun Cho, and Anna Korhonen. 2016.
\newblock \href {https://doi.org/10.18653/v1/n16-1162} {Learning distributed
  representations of sentences from unlabelled data}.
\newblock In \emph{{NAACL} {HLT} 2016, The 2016 Conference of the North
  American Chapter of the Association for Computational Linguistics: Human
  Language Technologies, San Diego California, USA, June 12-17, 2016}, pages
  1367--1377. The Association for Computational Linguistics.

\bibitem[{Hu and Liu(2004)}]{DBLP:conf/kdd/HuL04}
Minqing Hu and Bing Liu. 2004.
\newblock \href {https://doi.org/10.1145/1014052.1014073} {Mining and
  summarizing customer reviews}.
\newblock In \emph{Proceedings of the Tenth {ACM} {SIGKDD} International
  Conference on Knowledge Discovery and Data Mining, Seattle, Washington, USA,
  August 22-25, 2004}, pages 168--177. {ACM}.

\bibitem[{J{\"{a}}rvelin and
  Kek{\"{a}}l{\"{a}}inen(2002)}]{DBLP:journals/tois/JarvelinK02}
Kalervo J{\"{a}}rvelin and Jaana Kek{\"{a}}l{\"{a}}inen. 2002.
\newblock \href {https://doi.org/10.1145/582415.582418} {Cumulated gain-based
  evaluation of {IR} techniques}.
\newblock \emph{{ACM} Trans. Inf. Syst.}, 20(4):422--446.

\bibitem[{Kim et~al.(2021)Kim, Yoo, and Lee}]{DBLP:conf/acl/KimYL20}
Taeuk Kim, Kang~Min Yoo, and Sang{-}goo Lee. 2021.
\newblock \href {https://doi.org/10.18653/v1/2021.acl-long.197} {Self-guided
  contrastive learning for {BERT} sentence representations}.
\newblock In \emph{Proceedings of the 59th Annual Meeting of the Association
  for Computational Linguistics and the 11th International Joint Conference on
  Natural Language Processing, {ACL/IJCNLP} 2021, (Volume 1: Long Papers),
  Virtual Event, August 1-6, 2021}, pages 2528--2540. Association for
  Computational Linguistics.

\bibitem[{Kiros et~al.(2015)Kiros, Zhu, Salakhutdinov, Zemel, Urtasun,
  Torralba, and Fidler}]{DBLP:conf/nips/KirosZSZUTF15}
Ryan Kiros, Yukun Zhu, Ruslan Salakhutdinov, Richard~S. Zemel, Raquel Urtasun,
  Antonio Torralba, and Sanja Fidler. 2015.
\newblock \href
  {https://proceedings.neurips.cc/paper/2015/hash/f442d33fa06832082290ad8544a8da27-Abstract.html}
  {Skip-thought vectors}.
\newblock In \emph{Advances in Neural Information Processing Systems 28: Annual
  Conference on Neural Information Processing Systems 2015, December 7-12,
  2015, Montreal, Quebec, Canada}, pages 3294--3302.

\bibitem[{Le and Mikolov(2014)}]{DBLP:conf/icml/LeM14}
Quoc~V. Le and Tom{\'{a}}s Mikolov. 2014.
\newblock \href {http://proceedings.mlr.press/v32/le14.html} {Distributed
  representations of sentences and documents}.
\newblock In \emph{Proceedings of the 31th International Conference on Machine
  Learning, {ICML} 2014, Beijing, China, 21-26 June 2014}, volume~32 of
  \emph{{JMLR} Workshop and Conference Proceedings}, pages 1188--1196.
  JMLR.org.

\bibitem[{Li et~al.(2020)Li, Zhou, He, Wang, Yang, and
  Li}]{DBLP:conf/emnlp/LiZHWYL20}
Bohan Li, Hao Zhou, Junxian He, Mingxuan Wang, Yiming Yang, and Lei Li. 2020.
\newblock \href {https://doi.org/10.18653/v1/2020.emnlp-main.733} {On the
  sentence embeddings from pre-trained language models}.
\newblock In \emph{Proceedings of the 2020 Conference on Empirical Methods in
  Natural Language Processing, {EMNLP} 2020, Online, November 16-20, 2020},
  pages 9119--9130. Association for Computational Linguistics.

\bibitem[{Li et~al.(2007)Li, Burges, and Wu}]{DBLP:conf/nips/LiBW07}
Ping Li, Christopher J.~C. Burges, and Qiang Wu. 2007.
\newblock \href
  {https://proceedings.neurips.cc/paper/2007/hash/b86e8d03fe992d1b0e19656875ee557c-Abstract.html}
  {Mcrank: Learning to rank using multiple classification and gradient
  boosting}.
\newblock In \emph{Advances in Neural Information Processing Systems 20,
  Proceedings of the Twenty-First Annual Conference on Neural Information
  Processing Systems, Vancouver, British Columbia, Canada, December 3-6, 2007},
  pages 897--904. Curran Associates, Inc.

\bibitem[{Liu et~al.(2019)Liu, Ott, Goyal, Du, Joshi, Chen, Levy, Lewis,
  Zettlemoyer, and Stoyanov}]{DBLP:journals/corr/abs-1907-11692}
Yinhan Liu, Myle Ott, Naman Goyal, Jingfei Du, Mandar Joshi, Danqi Chen, Omer
  Levy, Mike Lewis, Luke Zettlemoyer, and Veselin Stoyanov. 2019.
\newblock \href {http://arxiv.org/abs/1907.11692} {Roberta: {A} robustly
  optimized {BERT} pretraining approach}.
\newblock \emph{CoRR}, abs/1907.11692.

\bibitem[{Logeswaran and Lee(2018)}]{DBLP:conf/iclr/LogeswaranL18}
Lajanugen Logeswaran and Honglak Lee. 2018.
\newblock \href {https://openreview.net/forum?id=rJvJXZb0W} {An efficient
  framework for learning sentence representations}.
\newblock In \emph{6th International Conference on Learning Representations,
  {ICLR} 2018, Vancouver, BC, Canada, April 30 - May 3, 2018, Conference Track
  Proceedings}. OpenReview.net.

\bibitem[{Ma et~al.(2016)Ma, Zhang, and He}]{DBLP:conf/asist/MaZH16}
Shutian Ma, Chengzhi Zhang, and Daqing He. 2016.
\newblock \href {https://doi.org/10.1002/pra2.2016.14505301065} {Document
  representation methods for clustering bilingual documents}.
\newblock In \emph{Creating Knowledge, Enhancing Lives through Information {\&}
  Technology - Proceedings of the 2016 Annual Meeting of the Association for
  Information Science and Technology, {ASIST} 2016, Copenhagen, Denmark,
  October 14-18, 2016}, volume~53 of \emph{Proc. Assoc. Inf. Sci. Technol.},
  pages 1--10. Wiley.

\bibitem[{Marelli et~al.(2014)Marelli, Menini, Baroni, Bentivogli, Bernardi,
  and Zamparelli}]{DBLP:conf/lrec/MarelliMBBBZ14}
Marco Marelli, Stefano Menini, Marco Baroni, Luisa Bentivogli, Raffaella
  Bernardi, and Roberto Zamparelli. 2014.
\newblock \href
  {http://www.lrec-conf.org/proceedings/lrec2014/summaries/363.html} {A {SICK}
  cure for the evaluation of compositional distributional semantic models}.
\newblock In \emph{Proceedings of the Ninth International Conference on
  Language Resources and Evaluation, {LREC} 2014, Reykjavik, Iceland, May
  26-31, 2014}, pages 216--223. European Language Resources Association
  {(ELRA)}.

\bibitem[{Mikolov et~al.(2013)Mikolov, Sutskever, Chen, Corrado, and
  Dean}]{DBLP:conf/nips/MikolovSCCD13}
Tom{\'{a}}s Mikolov, Ilya Sutskever, Kai Chen, Gregory~S. Corrado, and Jeffrey
  Dean. 2013.
\newblock \href
  {https://proceedings.neurips.cc/paper/2013/hash/9aa42b31882ec039965f3c4923ce901b-Abstract.html}
  {Distributed representations of words and phrases and their
  compositionality}.
\newblock In \emph{Advances in Neural Information Processing Systems 26: 27th
  Annual Conference on Neural Information Processing Systems 2013. Proceedings
  of a meeting held December 5-8, 2013, Lake Tahoe, Nevada, United States},
  pages 3111--3119.

\bibitem[{Pang and Lee(2004)}]{DBLP:conf/acl/PangL04}
Bo~Pang and Lillian Lee. 2004.
\newblock \href {https://doi.org/10.3115/1218955.1218990} {A sentimental
  education: Sentiment analysis using subjectivity summarization based on
  minimum cuts}.
\newblock In \emph{Proceedings of the 42nd Annual Meeting of the Association
  for Computational Linguistics, 21-26 July, 2004, Barcelona, Spain}, pages
  271--278. {ACL}.

\bibitem[{Pang and Lee(2005)}]{DBLP:conf/acl/PangL05}
Bo~Pang and Lillian Lee. 2005.
\newblock \href {https://doi.org/10.3115/1219840.1219855} {Seeing stars:
  Exploiting class relationships for sentiment categorization with respect to
  rating scales}.
\newblock In \emph{{ACL} 2005, 43rd Annual Meeting of the Association for
  Computational Linguistics, Proceedings of the Conference, 25-30 June 2005,
  University of Michigan, {USA}}, pages 115--124. The Association for Computer
  Linguistics.

\bibitem[{Pennington et~al.(2014)Pennington, Socher, and
  Manning}]{DBLP:conf/emnlp/PenningtonSM14}
Jeffrey Pennington, Richard Socher, and Christopher~D. Manning. 2014.
\newblock \href {https://doi.org/10.3115/v1/d14-1162} {Glove: Global vectors
  for word representation}.
\newblock In \emph{Proceedings of the 2014 Conference on Empirical Methods in
  Natural Language Processing, {EMNLP} 2014, October 25-29, 2014, Doha, Qatar,
  {A} meeting of SIGDAT, a Special Interest Group of the {ACL}}, pages
  1532--1543. {ACL}.

\bibitem[{Pobrotyn and Bialobrzeski(2021)}]{DBLP:journals/corr/abs-2102-07831}
Przemyslaw Pobrotyn and Radoslaw Bialobrzeski. 2021.
\newblock \href {http://arxiv.org/abs/2102.07831} {Neuralndcg: Direct
  optimisation of a ranking metric via differentiable relaxation of sorting}.
\newblock \emph{CoRR}, abs/2102.07831.

\bibitem[{Reimers and Gurevych(2019)}]{DBLP:conf/emnlp/ReimersG19}
Nils Reimers and Iryna Gurevych. 2019.
\newblock \href {https://doi.org/10.18653/v1/D19-1410} {Sentence-bert: Sentence
  embeddings using siamese bert-networks}.
\newblock In \emph{Proceedings of the 2019 Conference on Empirical Methods in
  Natural Language Processing and the 9th International Joint Conference on
  Natural Language Processing, {EMNLP-IJCNLP} 2019, Hong Kong, China, November
  3-7, 2019}, pages 3980--3990. Association for Computational Linguistics.

\bibitem[{Socher et~al.(2013)Socher, Perelygin, Wu, Chuang, Manning, Ng, and
  Potts}]{DBLP:conf/emnlp/SocherPWCMNP13}
Richard Socher, Alex Perelygin, Jean Wu, Jason Chuang, Christopher~D. Manning,
  Andrew~Y. Ng, and Christopher Potts. 2013.
\newblock \href {https://aclanthology.org/D13-1170/} {Recursive deep models for
  semantic compositionality over a sentiment treebank}.
\newblock In \emph{Proceedings of the 2013 Conference on Empirical Methods in
  Natural Language Processing, {EMNLP} 2013, 18-21 October 2013, Grand Hyatt
  Seattle, Seattle, Washington, USA, {A} meeting of SIGDAT, a Special Interest
  Group of the {ACL}}, pages 1631--1642. {ACL}.

\bibitem[{Su et~al.(2021)Su, Cao, Liu, and
  Ou}]{DBLP:journals/corr/abs-2103-15316}
Jianlin Su, Jiarun Cao, Weijie Liu, and Yangyiwen Ou. 2021.
\newblock \href {http://arxiv.org/abs/2103.15316} {Whitening sentence
  representations for better semantics and faster retrieval}.
\newblock \emph{CoRR}, abs/2103.15316.

\bibitem[{van~den Oord et~al.(2018)van~den Oord, Li, and
  Vinyals}]{DBLP:journals/corr/abs-1807-03748}
A{\"{a}}ron van~den Oord, Yazhe Li, and Oriol Vinyals. 2018.
\newblock \href {http://arxiv.org/abs/1807.03748} {Representation learning with
  contrastive predictive coding}.
\newblock \emph{CoRR}, abs/1807.03748.

\bibitem[{Volkovs and Zemel(2009)}]{DBLP:conf/icml/VolkovsZ09}
Maksims Volkovs and Richard~S. Zemel. 2009.
\newblock \href {https://doi.org/10.1145/1553374.1553513} {Boltzrank: learning
  to maximize expected ranking gain}.
\newblock In \emph{Proceedings of the 26th Annual International Conference on
  Machine Learning, {ICML} 2009, Montreal, Quebec, Canada, June 14-18, 2009},
  volume 382 of \emph{{ACM} International Conference Proceeding Series}, pages
  1089--1096. {ACM}.

\bibitem[{Voorhees and Tice(2000)}]{DBLP:conf/sigir/VoorheesT00}
Ellen~M. Voorhees and Dawn~M. Tice. 2000.
\newblock \href {https://doi.org/10.1145/345508.345577} {Building a question
  answering test collection}.
\newblock In \emph{{SIGIR} 2000: Proceedings of the 23rd Annual International
  {ACM} {SIGIR} Conference on Research and Development in Information
  Retrieval, July 24-28, 2000, Athens, Greece}, pages 200--207. {ACM}.

\bibitem[{Wang et~al.(2021)Wang, Reimers, and
  Gurevych}]{DBLP:conf/emnlp/Wang0G21}
Kexin Wang, Nils Reimers, and Iryna Gurevych. 2021.
\newblock \href {https://doi.org/10.18653/v1/2021.findings-emnlp.59} {{TSDAE:}
  using transformer-based sequential denoising auto-encoderfor unsupervised
  sentence embedding learning}.
\newblock In \emph{Findings of the Association for Computational Linguistics:
  {EMNLP} 2021, Virtual Event / Punta Cana, Dominican Republic, 16-20 November,
  2021}, pages 671--688. Association for Computational Linguistics.

\bibitem[{Wang and Isola(2020)}]{DBLP:conf/icml/0001I20}
Tongzhou Wang and Phillip Isola. 2020.
\newblock \href {http://proceedings.mlr.press/v119/wang20k.html} {Understanding
  contrastive representation learning through alignment and uniformity on the
  hypersphere}.
\newblock In \emph{Proceedings of the 37th International Conference on Machine
  Learning, {ICML} 2020, 13-18 July 2020, Virtual Event}, volume 119 of
  \emph{Proceedings of Machine Learning Research}, pages 9929--9939. {PMLR}.

\bibitem[{Wiebe et~al.(2005)Wiebe, Wilson, and
  Cardie}]{DBLP:journals/lre/WiebeWC05}
Janyce Wiebe, Theresa Wilson, and Claire Cardie. 2005.
\newblock \href {https://doi.org/10.1007/s10579-005-7880-9} {Annotating
  expressions of opinions and emotions in language}.
\newblock \emph{Lang. Resour. Evaluation}, 39(2-3):165--210.

\bibitem[{Wu and Zhao(2022)}]{DBLP:conf/emnlp/WuZ22}
Bohong Wu and Hai Zhao. 2022.
\newblock \href {https://aclanthology.org/2022.emnlp-main.221} {Sentence
  representation learning with generative objective rather than contrastive
  objective}.
\newblock In \emph{Proceedings of the 2022 Conference on Empirical Methods in
  Natural Language Processing, {EMNLP} 2022, Abu Dhabi, United Arab Emirates,
  December 7-11, 2022}, pages 3356--3368. Association for Computational
  Linguistics.

\bibitem[{Wu et~al.(2022{\natexlab{a}})Wu, Tao, Shen, Xu, Geng, and
  Jiang}]{DBLP:journals/corr/abs-2201-12093}
Qiyu Wu, Chongyang Tao, Tao Shen, Can Xu, Xiubo Geng, and Daxin Jiang.
  2022{\natexlab{a}}.
\newblock \href {http://arxiv.org/abs/2201.12093} {{PCL:} peer-contrastive
  learning with diverse augmentations for unsupervised sentence embeddings}.
\newblock \emph{CoRR}, abs/2201.12093.

\bibitem[{Wu et~al.(2022{\natexlab{b}})Wu, Gao, Lin, Han, Wang, and
  Hu}]{DBLP:journals/corr/abs-2210-06432}
Xing Wu, Chaochen Gao, Zijia Lin, Jizhong Han, Zhongyuan Wang, and Songlin Hu.
  2022{\natexlab{b}}.
\newblock \href {https://doi.org/10.48550/arXiv.2210.06432} {Infocse:
  Information-aggregated contrastive learning of sentence embeddings}.
\newblock \emph{CoRR}, abs/2210.06432.

\bibitem[{Xia et~al.(2008)Xia, Liu, Wang, Zhang, and
  Li}]{DBLP:conf/icml/XiaLWZL08}
Fen Xia, Tie{-}Yan Liu, Jue Wang, Wensheng Zhang, and Hang Li. 2008.
\newblock \href {https://doi.org/10.1145/1390156.1390306} {Listwise approach to
  learning to rank: theory and algorithm}.
\newblock In \emph{Machine Learning, Proceedings of the Twenty-Fifth
  International Conference {(ICML} 2008), Helsinki, Finland, June 5-9, 2008},
  volume 307 of \emph{{ACM} International Conference Proceeding Series}, pages
  1192--1199. {ACM}.

\bibitem[{Yan et~al.(2021)Yan, Li, Wang, Zhang, Wu, and
  Xu}]{DBLP:conf/acl/YanLWZWX20}
Yuanmeng Yan, Rumei Li, Sirui Wang, Fuzheng Zhang, Wei Wu, and Weiran Xu. 2021.
\newblock \href {https://doi.org/10.18653/v1/2021.acl-long.393} {Consert: {A}
  contrastive framework for self-supervised sentence representation transfer}.
\newblock In \emph{Proceedings of the 59th Annual Meeting of the Association
  for Computational Linguistics and the 11th International Joint Conference on
  Natural Language Processing, {ACL/IJCNLP} 2021, (Volume 1: Long Papers),
  Virtual Event, August 1-6, 2021}, pages 5065--5075. Association for
  Computational Linguistics.

\bibitem[{Zhang et~al.(2021)Zhang, He, Liu, Bing, and
  Li}]{DBLP:conf/acl/0004HLB020}
Yan Zhang, Ruidan He, Zuozhu Liu, Lidong Bing, and Haizhou Li. 2021.
\newblock \href {https://doi.org/10.18653/v1/2021.acl-long.402} {Bootstrapped
  unsupervised sentence representation learning}.
\newblock In \emph{Proceedings of the 59th Annual Meeting of the Association
  for Computational Linguistics and the 11th International Joint Conference on
  Natural Language Processing, {ACL/IJCNLP} 2021, (Volume 1: Long Papers),
  Virtual Event, August 1-6, 2021}, pages 5168--5180. Association for
  Computational Linguistics.

\bibitem[{Zhang et~al.(2020)Zhang, He, Liu, Lim, and
  Bing}]{DBLP:conf/emnlp/ZhangHLLB20}
Yan Zhang, Ruidan He, Zuozhu Liu, Kwan~Hui Lim, and Lidong Bing. 2020.
\newblock \href {https://doi.org/10.18653/v1/2020.emnlp-main.124} {An
  unsupervised sentence embedding method by mutual information maximization}.
\newblock In \emph{Proceedings of the 2020 Conference on Empirical Methods in
  Natural Language Processing, {EMNLP} 2020, Online, November 16-20, 2020},
  pages 1601--1610. Association for Computational Linguistics.

\bibitem[{Zhang et~al.(2022)Zhang, Zhu, Wang, Xu, Li, and
  Zhao}]{DBLP:conf/acl/ZhangZWXLZ22}
Yuhao Zhang, Hongji Zhu, Yongliang Wang, Nan Xu, Xiaobo Li, and Binqiang Zhao.
  2022.
\newblock \href {https://doi.org/10.18653/v1/2022.acl-long.336} {A contrastive
  framework for learning sentence representations from pairwise and triple-wise
  perspective in angular space}.
\newblock In \emph{Proceedings of the 60th Annual Meeting of the Association
  for Computational Linguistics (Volume 1: Long Papers), {ACL} 2022, Dublin,
  Ireland, May 22-27, 2022}, pages 4892--4903. Association for Computational
  Linguistics.

\bibitem[{Zhou et~al.(2022)Zhou, Zhang, Zhao, and
  Wen}]{DBLP:conf/acl/ZhouZZW22}
Kun Zhou, Beichen Zhang, Xin Zhao, and Ji{-}Rong Wen. 2022.
\newblock \href {https://doi.org/10.18653/v1/2022.acl-long.423} {Debiased
  contrastive learning of unsupervised sentence representations}.
\newblock In \emph{Proceedings of the 60th Annual Meeting of the Association
  for Computational Linguistics (Volume 1: Long Papers), {ACL} 2022, Dublin,
  Ireland, May 22-27, 2022}, pages 6120--6130. Association for Computational
  Linguistics.

\end{thebibliography}
\bibliographystyle{acl_natbib}

\clearpage
\appendix
\section{Training Details} \label{ap:detail}
We implement all experiments with the deep learning framework PyTorch on a single NVIDIA Tesla A100 GPU (40GB memory). We carry out grid-search of learning rate $\in \{$2e-5, 3e-5$\}$ and temperatures $\tau_2,\tau_3 \in \{0.0125,0.025,0.05\}$, while setting batch size to 128, temperature $\tau_1$ to 0.05, $\alpha$ to 1/3, $\beta$ to 1, $\gamma$ to 1 and the rate of linear scheduling warm-up to 0.05 for all the experiments. We train our models for 4 epochs, and evaluate the model every 125 steps on the dev set of STS-B and keep the best checkpoint for the final evaluation on test sets of all STS and TR tasks. The hyperparameter settings we adopt are shown in Table \ref{tab:hyper}. Following SimCSE, we utilize the embedding corresponding to [CLS] token as the representation of the input sentence. We utilize SimCSE-BERT$_{\rm base}$ and SimCSE-BERT$_{\rm large}$ as a multi-teacher for \gmodelnospace-BERT$_{\rm base}$ and \gmodelnospace-BERT$_{\rm large}$, while SimCSE-RoBERTa$_{\rm base}$ and SimCSE-RoBERTa$_{\rm large}$ as a multi-teacher for \gmodelnospace-RoBERTa$_{\rm base}$ and \gmodelnospace-RoBERTa$_{\rm large}$.


\section{DiffCSE Settings for Transfer Tasks} \label{ap:tr}
DiffCSE uses different dev sets to find the best hyperparameters for the two tasks (STS-B dev set for STS tasks, dev sets of 7 TR tasks for TR tasks), while other methods only use the STS-B dev set for both tasks, which is not fair. Therefore we obtain the results in Table \ref{tab:tranr} from its publicly available code and checkpoints for STS tasks\footnote{\url{https://github.com/voidism/DiffCSE}} instead of directly importing the results from its original paper. For a more comprehensive comparison with DiffCSE on TR tasks, we also use dev sets of 7 TR tasks to find the best hyperparameters and checkpoints. As shown in Table \ref{tab:apptranr}, \gmodel still outperforms DiffCSE in this setting. 

\section{Data Statistics}
\label{ap:statistics}
The complete listings of train/dev/test stats of STS and TR datasets can be found in Table \ref{tab:sts_stats} and \ref{tab:tr_stats}, respectively. Note that for STS tasks, we only use test sets for the final evaluation and dev set of STS-B to find best hyperparameters and checkpoints. The train sets of all STS datasets are not used in our experiments. For TR tasks, we follow the default settings of SentEval toolkit \cite{DBLP:conf/lrec/ConneauK18} to use 10-fold evaluation for all TR datasets except SST. We can directly use the already split datasets to evaluate on SST.

\begin{table}[t]
\centering
\small
	\begin{tabular}{cccc}
    \toprule
    Dataset & Train & Dev & Test \\
    \midrule
    STS12 & - & - & 3108 \\ 
    STS13 & - & - & 1500 \\
    STS14 & - & - & 3750 \\
    STS15 & - & - & 3000 \\
    STS16 & - & - & 1186 \\
    STS-B & 5749 & 1500 & 1379 \\
    SICK-R & 4500 & 500 & 4927 \\
    \bottomrule
    \end{tabular}
    \vspace{-2mm}
    \caption{A listing of train/dev/test stats of STS datasets.}
    \label{tab:sts_stats}
    \vspace{-3mm}
\end{table}

\begin{table}[t]
\centering
\small
    \begin{tabular}{cccc}
    \toprule
    Dataset & Train & Dev & Test \\
    \midrule
    MR & 10662 & - & - \\ 
    CR & 3775 & - & - \\
    SUBJ & 10000 & - & - \\
    MPQA & 10606 & - & - \\
    SST & 67349 & 872 & 1821 \\
    TREC & 5452 & - & 500 \\
    MRPC & 4076 & - & 1725 \\
    \bottomrule
    \end{tabular}
    \vspace{-2mm}
    \caption{A listing of train/dev/test stats of TR datasets.}
    \label{tab:tr_stats}
    \vspace{-3mm}
\end{table}
\begin{table*}[t]
\centering
\small
\begin{adjustbox}{width=1.7\columnwidth,center}
  \begin{tabular}{c|cccccccc}
    \toprule
     & \multicolumn{4}{c}{\bf \gmodelnospace-BERT} & \multicolumn{4}{c}{\bf \gmodelnospace-RoBERTa} \\
     & \multicolumn{2}{c}{base} & \multicolumn{2}{c}{large} & \multicolumn{2}{c}{base} & \multicolumn{2}{c}{large} \\
     & listNet & listMLE & listNet & listMLE & listNet & listMLE & listNet & listMLE \\
     \midrule
     Batch size & 128 & 128 & 128 & 128 & 128 & 128 & 128 & 128 \\
     Learning rate & 3e-5 & 2e-5 & 3e-5 & 2e-5 & 2e-5 & 3e-5 & 3e-5 & 3e-5 \\
     $\tau_1$ & 0.05 & 0.05 & 0.05 & 0.05 & 0.05 & 0.05 & 0.05 & 0.05 \\
     $\tau_2$ & 0.025 & 0.05 & 0.05 & 0.05 & 0.05& 0.05& 0.025 & 0.025 \\ 
     $\tau_3$ & 0.0125 & - & 0.025 & -  & 0.025 & - & 0.0125 & - \\ 
     \bottomrule
\end{tabular}
\end{adjustbox}
\vspace{-2mm}
\caption{The hyperparameter values for \gmodel training.}
\label{tab:hyper}
\vspace{-3mm}
\end{table*}
\begin{table*}[t]
\small
\centering
\begin{adjustbox}{width=1.9\columnwidth,center}
  \begin{tabular}{p{2cm}|l|cccccccc}
    \toprule
    \bf PLMs & \bf Methods  & \bf MR & \bf CR & \bf SUBJ & \bf MPQA & \bf SST & \bf TREC & \bf MRPC & \bf avg. \\
    \midrule
    \multirow{3}{*}{BERT$_{\rm base}$} & +DiffCSE & 82.69 & 87.23 & \underline{95.23} & 89.28 & 86.60 & 90.40 & 76.58 & 86.86 \\
    & \bf +\gmodelnospace$_{\rm listNet}$ & \bf 83.64 & \bf 88.32 & \bf 95.26 & \underline{89.99} &  \bf 89.02 & \bf 90.80 & \bf 77.10 & \bf 87.73 \\
    & \bf +\gmodelnospace$_{\rm listMLE}$ & \underline{83.05} & \underline{88.03} & 95.13 & \bf 90.00 & \underline{88.41} & \underline{90.60} & \underline{76.81} & \underline{87.43} \\
    \midrule
    \multirow{3}{*}{RoBERTa$_{\rm base}$} & +DiffCSE  & 82.82 & 88.61 & \bf{94.32} & 87.71 & 88.63 & 90.40 & \underline{76.81} & 87.04\\
    & \bf +\gmodelnospace$_{\rm listNet}$ & \bf 83.84 & \underline{88.93} & \underline{94.21} & \underline{89.17} & \bf 90.23 & \bf 91.60 & \bf 77.28 & \bf 87.89 \\
    & \bf +\gmodelnospace$_{\rm listMLE}$ & \underline{83.38} & \bf 89.04 & 94.17 & \bf 89.23 & \underline{89.51} & \underline{91.40} & 76.58 & \underline{87.62} \\
    \bottomrule
\end{tabular}
\end{adjustbox}
\vspace{-2mm}
\caption{Sentence representations performance on TR tasks (accuracy) using the dev sets of 7 TR tasks to find the best hyperparameters. The results of DiffCSE are from its original paper. We mark the best (bold) and second-best (underlined) results among models with the same PLMs.}
\label{tab:apptranr}
\vspace{-3mm}
\end{table*}

\begin{table}[t]
\centering
\small
\begin{adjustbox}{width=0.95\columnwidth,center}
  \begin{tabular}{c|cccc}
    \toprule
    & \multicolumn{2}{c}{\bf SimCSE} & \multicolumn{2}{c}{\bf \gmodel} \\
    & base & large & base & large \\
    \midrule
    Batch size & 128 & 128 & 128 & 128 \\
    Epoch & 1 & 1 & 4 & 4 \\
    Time & 20min & 45min  & 120min & 220min\\
    Time per epoch & 20min & 45min & 30min & 55min \\
    \bottomrule
\end{tabular}
\end{adjustbox}
\caption{Training efficiency of SimCSE and \gmodelnospace. SimCSE$_{\rm base}$ and SimCSE$_{\rm large}$ provide pseudo ranking labels for every \gmodel model.}
\label{tab:efficiency}
\vspace{-3mm}
\end{table}
\section{Training Efficiency}
\label{ap:efficiency}
We compare the training efficiency of SimCSE and \gmodel, which are tested on a single NVIDIA Tesla A100 GPU (40GB memory). We set batch size to 128 for both SimCSE and \gmodelnospace, and training epoch to their original settings (1 for SimCSE, 4 for \gmodelnospace). \gmodel utilizes SimCSE$_{\rm base}$ and SimCSE$_{\rm large}$ as a multi-teacher to provide pseudo ranking labels. As shown in Table \ref{tab:efficiency}, \gmodelnospace$_{\rm base}$ and \gmodelnospace$_{\rm large}$ can be trained within 2 hours and 3.7 hours respectively. Since \gmodel needs to calculate pseudo ranking labels of the teacher, it requires additional training time per epoch than SimCSE.

\begin{figure*}[tbp]
\centering
\subfloat[SimCSE]{\includegraphics[width=0.32\textwidth]{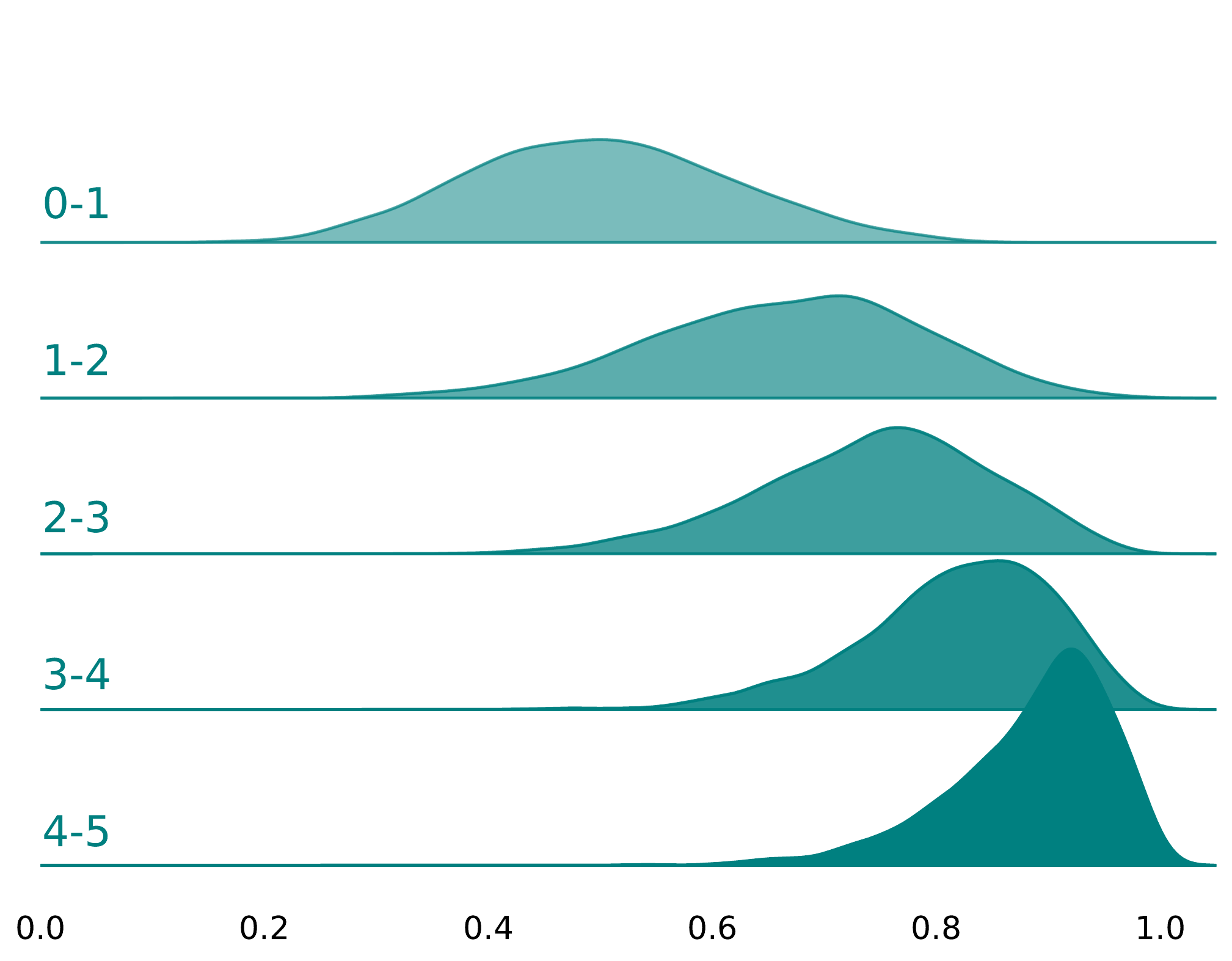}\label{fig:simcse_dis}}
\subfloat[DiffCSE]{\includegraphics[width=0.32\textwidth]{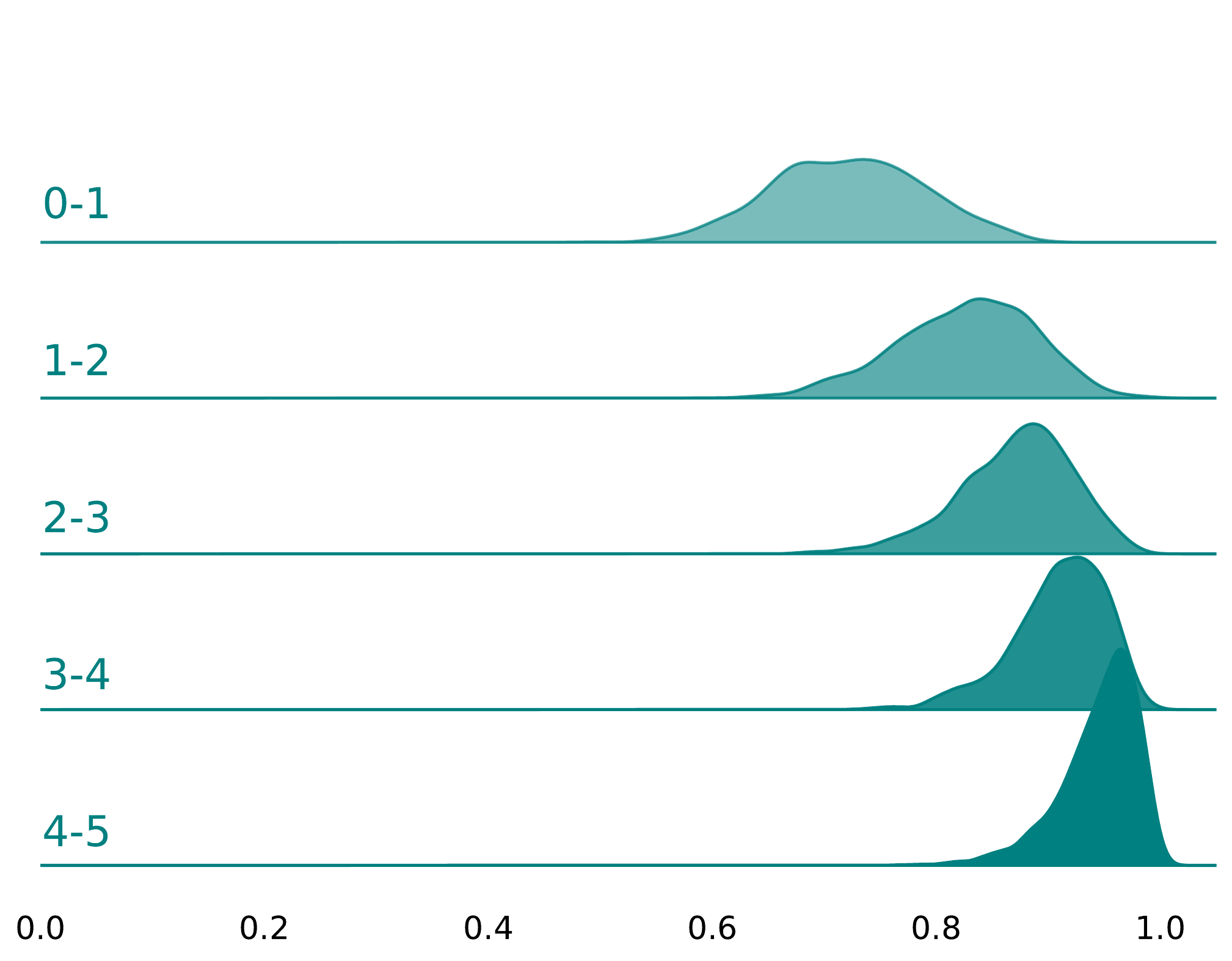}\label{fig:diffcse_dis}}
\subfloat[\gmodel]{\includegraphics[width=0.32\textwidth]{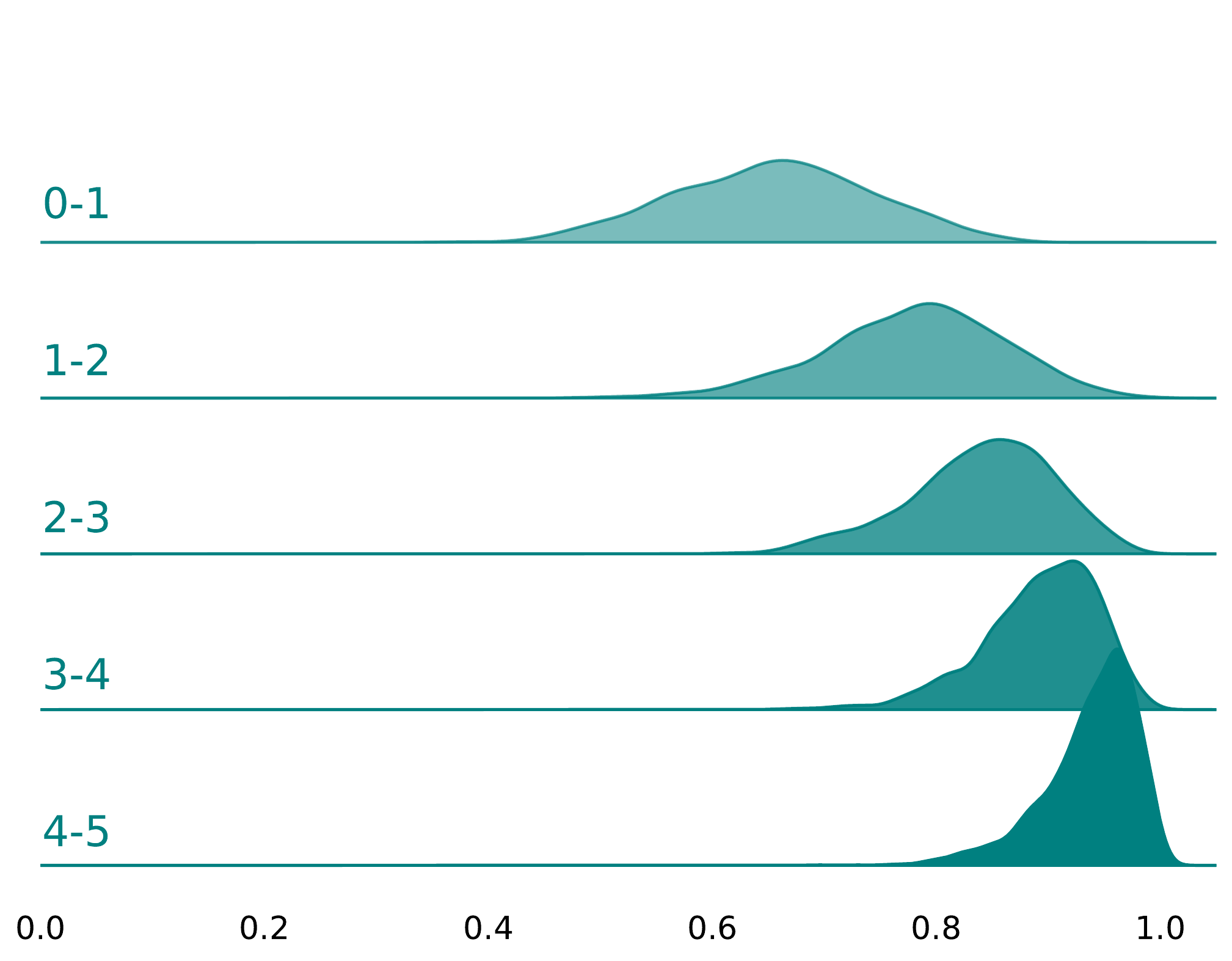}\label{fig:listmle_dis}}
\caption{The distribution of cosine similarity for sentence pairs of STS-B dev set. Along the y-axis are 5 groups of pairs split based on ground truth ratings, and x-axis is the cosine similarity.}
\label{fig:dis}
\end{figure*}

\begin{table*}[t]
\begin{adjustbox}{width=1.95\columnwidth,center}
\small
\begin{tabular}{p{9.5cm}ccc}
\toprule
\makecell[c]{\bf Target Sentences} & \bf Label & \bf SimCSE & \bf \gmodel \\
\midrule
$\bullet$ a and c are on the same closed path with the battery & 3.60 (1) & 0.81 (1) & 0.90 (1)\\
$\bullet$ bulb a and bulb c affect each other. & 2.80 (2) & 0.58 (3) & 0.75 (2) \\
$\bullet$ the are on the same wire & 1.60 (3) & 0.60 (2) & 0.68 (3) \\
$\bullet$ because breaking one bulb then affects the ability of the others to light up. & 1.20 (4) & 0.37 (5) & 0.59 (4)\\
$\bullet$ if one bulb is removed , the others stop working & 0.60 (5) & 0.38 (4) & 0.54 (5) \\
\midrule
\multicolumn{4}{l}{\textbf{Query Sentence:} a and c are in the same closed path} \\
\midrule
\midrule
$\bullet$ because by measuring voltage, you find the gap where there's a difference in electrical states.  & 3.80 (1) & 0.86 (1) & 0.90 (1)\\
$\bullet$ it allows you to measure electrical states between terminals & 3.20 (2) & 0.64 (3) & 0.84 (2) \\
$\bullet$ it checks the electrical state between two terminals. & 2.60 (3) & 0.65 (2) & 0.78 (3) \\
$\bullet$ find where there are different electrical states & 2.60 (3) & 0.55 (5) & 0.78 (3)\\
$\bullet$ you can see where the gap is & 2.20 (5) & 0.62 (4) & 0.69 (5) \\
\midrule
\multicolumn{4}{l}{\textbf{Query Sentence:} measuring voltage indicates the place where the electrical state changes due to a gap.} \\
\bottomrule
\end{tabular}
\end{adjustbox}
\captionof{table}{Two examples of a query sentence and several target sentences from the STS datasets, with their similarity scores and rankings. The label scores are from human annotations. The SimCSE and \gmodel similarity scores are from the model predictions respectively, with the corresponding ranking positions. It can be seen that sentence rankings based on SimCSE are incorrect, while \gmodel generates more effective scores with accurate rankings.}
\label{tab:appendix_case}
\end{table*}

\section{Cosine Similarity Distribution}
\label{ap:distribution}
We demonstrate the distribution of cosine similarities for sentence pairs of STS-B dev set in Figure \ref{fig:dis}. We can observe that cosine similarity distributions from all models are consistent with human ratings. However, the cosine similarities of \gmodel are slightly higher than that of SimCSE under the same human rating, as \gmodel pulls similar negatives closer during incorporating ranking consistency and ranking distillation, and shows lower variance. Compared with DiffCSE, \gmodel shows a more scattered distribution. This observation further validates that \gmodel can achieve a better alignment-uniformity balance.

\section{Case Study} \label{ap:case}
We present another two examples of a query sentence and several target sentences from the STS datasets, with their similarity scores and rankings in Table \ref{tab:appendix_case}. It is obvious that the similarity scores produced by \gmodel are more effective than SimCSE, with consistent rankings to the ground-truth labels. It further demonstrates that SimCSE only captures coarse-grained semantic ranking information via contrastive learning, while \gmodel can capture fine-grained semantic ranking information. For example, SimCSE can distinguish between similar and dissimilar sentences, however, it can not distinguish between very similar and less similar sentences as RankCSE.   

\begin{table*}[t]
\small
\centering
\begin{adjustbox}{width=1.95\columnwidth,center}
  \begin{tabular}{c|l|cccccccc}
    \toprule
    \bf Metrics & \bf Methods  & \bf STS12 & \bf STS13 & \bf STS14 & \bf STS15 & \bf STS16 & \bf STS-B & \bf SICK-R & \bf avg. \\
    \midrule
    \multirow{3}{*}{KCC} & +SimCSE & 36.08 & 36.60 & \underline{44.14} & 49.02 & 54.66 & 58.44 & \underline{54.65} & 47.66\\
    & +DiffCSE & \underline{38.59} & \underline{41.89} & 42.37 & \underline{51.19} & \bf 58.90 & \underline{59.21} & 53.42 & \underline{49.37}\\
    & \bf +\gmodelnospace & \bf 42.79 & \bf 46.26 & \bf 44.53 & \bf 52.00 & \underline{57.21} & \bf 63.64 & \bf 57.40 & \bf 51.98 \\
    \midrule
    \multirow{3}{*}{NDCG} & +SimCSE & 97.80 & 89.33 & 92.71 & \underline{96.93} & 94.28 & 96.49 & \underline{98.44} & 95.14 \\
    & +DiffCSE & \bf 98.35 & \underline{90.22} & \underline{93.05} & 96.91 & \underline{94.79} & \underline{97.05} & 98.34 & \underline{95.53} \\
    & \bf +\gmodelnospace & \underline{98.20} & \bf 92.27 & \bf 93.46 & \bf 97.21 & \bf 95.24 & \bf 97.45 & \bf 98.67 & \bf 96.07 \\
    \bottomrule
\end{tabular}
\end{adjustbox}
\caption{Sentence representations performance on ranking tasks (KCC and NDCG) using BERT$_{\rm base}$. The results of SimCSE and DiffCSE are obtained from their publicly available codes and checkpoints. We mark the best (bold) and second-best (underlined) results.}
\label{tab:rank_metrics}
\end{table*}

\section{Ranking Tasks} \label{ap:rank}
We build the ranking task based on each STS dataset to verify that \gmodel can capture fine-grained semantic ranking information. For one sentence $x_i$, if there are more than three sentence pairs $(x_i,x_i^j)$ containing $x_i$ with similarity score label $y_i^j$ in the dataset, we view $\{x_i,x_i^j,y_i^j\}_{j=1}^k (k > 3)$ as a sample of the ranking task, as shown in Table \ref{tab:appendix_case}. We adopt KCC (Kendall’s correlation coefficient \cite{abdi2007kendall}) and NDCG  (normalized discounted cumulative gain \cite{DBLP:journals/tois/JarvelinK02}) as evaluation metrics for ranking tasks, and demonstrate the results in Table \ref{tab:rank_metrics}. \gmodel outperforms SimCSE and DiffCSE on both KCC and NDCG, which validates that \gmodel can capture fine-grained semantic ranking information by incorporating ranking consistency and ranking distillation. Another observation is that SimCSE and DiffCSE also achieve moderate results, which shows they can distinguish coarse-grained semantic differences via contrastive learning.  

\section{Alignment and Uniformity}\label{ap:aligment-uniformity}
\citet{DBLP:conf/icml/0001I20} use two properties related to contrastive learning, alignment and uniformity, to measure the quality of representation space. Alignment calculates expected distance between normalized representations of positive pairs $p_{\rm{pos}}$:
\begin{equation}
    \small
    \label{eq:alignment}
    \ell_{\rm{align}}\triangleq \underset{(x, x^+)\sim p_{\rm{pos}}}{\mathbb{E}} \Vert f(x) - f(x^+) \Vert^2,
\end{equation}
while uniformity measures how well the normalized representations are uniformly distributed:
\begin{equation}
    \small
    \label{eq:uniformity}
    \ell_{\rm{uniform}}\triangleq\log \underset{~~~x, y\stackrel{i.i.d.}{\sim} p_{\rm{data}}}{\mathbb{E}}   e^{-2\Vert f(x)-f(y) \Vert^2},
\end{equation} 
where $p_{\rm{data}}$ denotes the distribution of sentence pairs. Smaller alignment means positive instances have been pulled closer, while smaller uniformity means random instances scatter on the hypersphere. These two measures are smaller the better, and well aligned with the object of contrastive learning.

\end{document}